\documentclass[runningheads]{llncs}
\usepackage[T1]{fontenc}
\usepackage{graphicx}
\usepackage{amssymb}
\usepackage{amsmath}
\usepackage{amsmath}
\usepackage{mathtools}
\usepackage[table,xcdraw]{xcolor}
\usepackage{multirow}
\usepackage{lineno}

\usepackage{xcolor}
\usepackage{enumitem}
\usepackage{adjustbox}

\newcommand{\lb}[1]{{\color{black}#1}}
\newcommand{\np}[1]{{\color{black} #1}}
\newcommand{\fc}[1]{{\color{black}#1}}
\newcommand{\fz}[1]{{\color{black}#1}}

\newcommand{\target}{y}
\newcommand{\cond}{x}
\newcommand{\fullseries}{\overline{x}}
\newcommand{\fulldataset}{\mathcal{D}}
\newcommand{\datasetTrain}{\mathcal{D}_T}
\newcommand{\datasetCP}{\mathcal{D}_{CP}}
\newcommand{\datasetAD}{\mathcal{D}_{AD}}
\newcommand{\quantileLow}[1]{\hat{q}_{l}({#1})}
\newcommand{\quantileHigh}[1]{\hat{q}_{u}({#1})}
\newcommand{\CquantileLow}[1]{\hat{q}^c_{l}({#1})}
\newcommand{\CquantileHigh}[1]{\hat{q}^c_{u}({#1})}

\newcommand{\deltaSeq}[2]{\Delta^{#1}_{#2}}
\newcommand{\deltaObs}[3]{\delta^{#1}_{{#2}, {#3}}} 
\newcommand{\deltaFit}[3]{\hat{\delta}^{#1}_{{#2}, {#3}}}

\newcommand{\betaObs}[3]{\beta^{#1}_{{#2}, {#3}}} 
\newcommand{\betaSeq}[2]{B^{#1}_{#2}}

\newcommand{\uSeq}[2]{U^{#1}_{#2}}
\newcommand{\uObs}[3]{u^{#1}_{{#2}, {#3}}} 

\newcommand{\zSeq}[2]{Z^{#1}_{#2}}
\newcommand{\zObs}[3]{z^{#1}_{{#2}, {#3}}} 

\newcommand{\SigmaFit}{\hat{\Sigma}_j}

\newcommand{\ncf}{\mathtt{ncf}}

\newcommand{\estEDF}[3]{\hat{F}_{{#1}, {#2}}\left({#3}\right)}

\newcommand{\estSEDF}[3]{\hat{F}_{{#1}, {#2}}({#3})}

\newcommand{\mahaDist}[1]{D_M({#1})}
\newcommand{\mahaDistSquare}[1]{D^2_M({#1})}

\begin{document}
%
\title{CoCAI: Copula-based Conformal Anomaly Identification for Multivariate Time-Series}
%
\titlerunning{CoCAI}
%
\author{Nicholas A. Pearson\inst{1}
\and
 Francesca Zanello \inst{2} \and \newline Davide Russo\inst{2} \and 
Luca Bortolussi\inst{1}
\and 
Francesca Cairoli\inst{1}
}
\authorrunning{N. A. Pearson et al.}
%
\institute{
University of Trieste, Trieste, Italy\newline 
\and Idrostudi srl, Trieste, Italy
}

\maketitle              
%

\begin{abstract}
We propose a novel framework that harnesses the power of generative artificial intelligence and copula-based modeling to address two critical challenges in multivariate time-series analysis: delivering accurate predictions and enabling robust anomaly detection. Our method, Copula-based Conformal Anomaly Identification for Multivariate Time-Series (CoCAI), leverages a diffusion-based model to capture complex dependencies within the data, enabling high quality forecasting. The model's outputs are further calibrated using a conformal prediction technique, yielding predictive regions which are statistically valid, i.e., cover the true target values with a desired confidence level. Starting from these calibrated forecasts, robust outlier detection is performed by combining dimensionality reduction techniques with copula-based modeling, providing a statistically grounded anomaly score. CoCAI benefits from an offline calibration phase that allows for minimal overhead during deployment and delivers actionable results rooted in established theoretical foundations. Empirical tests conducted on real operational data derived from water distribution and sewerage systems confirm CoCAI's effectiveness in accurately forecasting target sequences of data and in identifying anomalous segments within them.

\end{abstract}

\section{Introduction}\label{sec:intro}
\np{Recent advances in sensor, network and storage technology have expanded opportunities for critical system monitoring and control, while also reducing data collection costs~\cite{ahmed2023industrialMonitoring}.} As a result, such campaigns have been carried out in a variety of critical domains, from urban and environmental engineering~\cite{lanzolla2021EnvironmentalMonitoring,zanella2014SmartCitiesMonitoring,rashid2016urbanSensorNetworksMonitoring} to industrial applications~\cite{wright2019corrosionMonitoring,ren2021bigIndustrialMonitoring}, healthcare~\cite{ketu2021HealthMonitoring,majumder2017OldPeopleMonitoring,taylor2020HEalthMonitoring} and finance~\cite{mhlanga2024FinanceMonitoring}. The data recorded over time by these systems allows for the creation of an ordered sequence of multiple interrelated values, commonly referred to as a multivariate time-series (MTS). These time-series play an essential role in monitoring the correct behavior of the system as they offer a near-real time representation of its state. These data can further facilitate the implementation of predictive alert systems designed to detect critical events at an early stage, enabling timely and targeted mitigating measures that reduce their operational costs and the likelihood of major failures~\cite{hu2022integrated}. In most fields of application, the sensors responsible for data collection are placed in challenging and uncontrolled locations, increasing the risk of environmental and operational damage, interference or total failure. If these factors are overlooked, the data recorded by these sensors may be corrupted or anomalous, yielding readings that do not fully reflect the system's true state.
Undetected anomalous readings undermine the monitoring system's reliability, potentially generating false alarms or, more critically, failing to identify high-risk events. This degradation directly compromises the integrity of downstream data-driven decision-making processes.
\fz{From a theoretical standpoint, an anomaly  
 can be defined as }``\emph{an observation that deviates so much from other observations as to arouse suspicion that is was generated by a different mechanism}'' \cite{hawkins1980identification}. More specifically, in the domain of MTS, anomalies may take the form of punctual observations or sequences of points which present patterns and characteristics which fail to conform to the expected behavior \cite{chandola2009anomaly} and the available historic data. Anomalies and outliers in time-series data generally fall into two categories: 
 measurement errors/noise or 
 system events like failures or operational changes \cite{boniol2024dive}. 
An accurate and timely identification of both 
is crucial. 
In the case of noise, removing such artifacts ensures data integrity.
For system events, early detection offers valuable diagnostic information, enabling data-driven interventions that can prevent operational disruptions or optimize system performance.\newline  \indent We propose CoCAI (Copula-based Conformal Anomaly Identification), a novel framework for multivariate time-series which allows for both 
 a statistically valid imputation of any missing segment of the time-series 
 and for the detection of anomalous patterns via a copula-based modeling approach. 
 \np{CoCAI's predictive component utilizes a score based diffusion model specifically designed for time-series data \cite{tashiro2021csdi}} to generate imputations for any target segment of the multivariate sequence. To provide rigorous coverage guarantees, these imputations are then adjusted using a conformal prediction \np{technique} tailored for time series~\cite{sun2023copulaCPTS}. After the predictive step, CoCAI computes 
 deviations between predicted and observed values using an suitable distance metric, yielding a sequence 
 that quantifies the discrepancy from the predicted behavior.
 This temporal sequence of discrepancies is then compressed via a dimensionality reduction technique that preserves temporal dependencies. By applying this procedure to a held-out dataset, we derive empirical univariate distributions for each term in the low-dimensional representation. These distributions form the basis for estimating their joint distribution using copula modeling. Building on established theoretical results, including the Mahalanobis distance and the inferred joint distribution, CoCAI produces an intuitive, interpretable anomaly score.
 
\noindent The \emph{main advantages} of the proposed method are:
 \begin{enumerate}[label=\roman*.]
    
     \item CoCAI provides accurate imputations for target data by leveraging score-based diffusion models in combination with conformal prediction, ensuring both high accuracy and statistically valid coverage guarantees.
     \item CoCAI offers significant flexibility, as imputations can be performed and calibrated across highly variable target windows, allowing it to adapt to different use cases and requirements.
     \item CoCAI provides a statistically sound anomaly score, allowing for flexibility in threshold selection and for ease of interpretation of the results. Furthermore, the anomaly detection step remains agnostic to 
     the selected predictive model.
     \item Given the use of dimensionality reduction techniques, CoCAI 
     scales well to longer forecasting windows. 
     \item At runtime, CoCAI introduces minimal computational overhead and can be deployed instantaneously after a one-time offline calibration.
 \end{enumerate}

 \noindent \textbf{Related Works.}
\emph{Conformal prediction} has been widely applied to calibrate the output of runtime monitoring tasks~\cite{cairoli2023learning,cairoli2025conformal,cairoli2023conformal,lindemann2023conformal,zhao2024robust,cairoli2025scalable}. Our work directly conforms the temporal outputs rather than a scalar assurance metric.
\emph{Anomaly detection} in time series has been explored via next-step prediction~\cite{russo2025robust} and statistical methods, including majority-voting in ensembles~\cite{torfah2023learning}, copula-based approaches~\cite{li2020copod} and variational techniques~\cite{wang2022variational}. Surveys~\cite{schmidl2022anomaly,liu2024elephant} and domain-specific criteria~\cite{bertrand2021metrology} further contextualize the challenge of detecting distributional shifts.
\vspace{2mm}


\noindent Building on Section \ref{sec:background}, we introduce the full CoCAI methodology in Section \ref{sec:methodology} and validate it through two real-world case studies (water distribution and sewerage systems) in Section \ref{sec:experiments}.


\section{Background}
\label{sec:background}

In this section, we first present the predictive model used for time-series imputation -- a score-based diffusion model -- along with the conformal inference framework 
which allows for the calibration of its predictions into statistically valid regions. We then introduce copulas and their role in modeling temporal dependencies, which is crucial for time series analysis in both calibration and anomaly detection. \np{Lastly, we provide a definition of the Mahalanobis distance and summarize its properties under known distributions.} While diffusion models were selected for their flexibility and robustness, our method is general and can be adapted to other predictive models.

\subsection{Diffusion Models for Time Series Imputation}\label{subsec:DDPMs}
As probabilistic models, generative models learn to estimate the underlying data distribution from samples while preserving key statistical characteristics.
Once trained, these models can be used to generate new synthetic observations which resemble the training data. 
Sampling from the learned probabilistic model involves drawing from a base distribution and transforming it into the desired output. Here, the neural network serves as a distribution transformer.
Denoising Diffusion Probabilistic Models (DDPMs) \cite{hoDDPM} are a type of deep generative models that have been shown to produce high quality samples in a variety of fields. 
DDPMs are trained in a two step procedure to learn an approximation of a true data distribution. In the initial forward process, data is gradually and iteratively corrupted by adding small amounts of Gaussian noise via a Markov chain ($\mathcal{K}$ iterative diffusion steps). In the complementary backwards process, a neural network is trained to gradually estimate and remove the noise at each step, learning an iterative procedure that allows the generation of clean samples starting from pure random noise. More formally, let $x_0$ be data belonging to the unknown target data distribution $q(x_0)$, and let $p_\theta(x_0)$ be a model distribution that aims to approximate $q(x_0)$.
As $k$ approaches $\mathcal{K}$, the distribution of $x_k$ converges to a standard Gaussian distribution. 
Starting from pure random noise, a data sample is generated by iteratively sampling from the reverse transition kernel $p_{\theta}\left(x_{k-1}\mid x_{k}\right)$ until $k=1$. The learnable transition kernel introduced in~\cite{hoDDPM} takes the form:
    $p_{\theta}\left(x_{k-1} \mid x_k \right) = \mathcal{N}\left( \mu_{\theta}(x_k, k), \sigma_\theta\left(x_k, k\right)I\right),$
where the variance $\sigma_{\theta}\left(x_k, k\right)$ derives from the constant rates used during the noise injection phase. 
The mean parameter $\mu_{\theta}\left(x_k, k\right)$ 
also depends on a learnable denoising function $\epsilon_{\theta}\left(x_k, k\right)$, usually in the form of a neural network, that estimates the noise added to $x_k$.
With their work on Conditional Score-based Diffusion Models for Probabilistic Time Series Imputation (CSDI) \cite{tashiro2021csdi}, Tashiro et al. adapted DDPMs specifically for the task of time-series imputation. Let $x\in \mathbb{R}^{D\times L}$ be a multivariate time-series with missing values, where $D$ and $L$ represent the numbers of channels and timesteps respectively. It is possible to identify the missing values of $x$ using a binary mask $M\in\{0,1\}^{D\times L}$, where $M_{i,j}=0$ if the corresponding element of $x$ is missing. The task of time-series imputation focuses on estimating these missing values given the remaining observed values of the time-series. 
For a more rigorous and detailed treatment of DDPMs and their adaptation in the CSDI framework, we refer the reader to Section~\ref{app:ddpm} in Appendix~\ref{app:bg}.


\subsection{Conformal Prediction}
\label{subsec:conformalPrediction}
Given the growth in popularity of predictive models in high-risk scenarios, it is often important not only to obtain an accurate forecast for the values of interest but also an estimate of the uncertainty associated with those predictions \cite{introToCP}, in other words, a measure of their reliability. 
Uncertainty can be estimated via \emph{conformal prediction} (CP), a framework that constructs predictive regions 
that offer statistically sound guarantees of covering the ground truth with a desired probability. CP is flexible enough to 
produce these regions 
starting from the output of most 
predictive models, 
requiring minimal assumptions, namely exchangeability, on the data.
We here focus on regression problems. Let $\mathcal{I}=\{(x_i, y_i)\}_{i=1}^n$ be a dataset of exchangeable samples, where $x_i \in \mathcal{X}$ represents input variables and $y_i \in \mathcal{Y}$ denotes the corresponding target value. In \emph{split CP}~\cite{introToCP} we divide $\mathcal{I}$ into a training dataset $\mathcal{I}_{T} =\{(x_i, y_i)\}_{i=1}^m$ and a calibration set $\mathcal{I}_{C} =\{(x_i, y_i)\}_{i=m+1}^n$. Let $\hat{f}$ be a predictive model fitted on the training dataset. CP revolves around a notion of nonconformity defined by a function that quantifies the discrepancy between the values predicted by the model and the ground truth. 
Let $\ncf:\mathcal{X}\times \mathcal{Y} \to \mathbb{R}$ be the selected non-conformity function (NCF) and let $\mathcal{S}_C$ be the set of non-conformity scores $s_i =\ncf (x_i, y_i)$  for all samples of the calibration set $\mathcal{I}_{C}$.
Given an error 
probability $\alpha\in (0, 1)$, under the assumption that $(x,y), (x_{m+1},y_{m+1}),\dots,(x_{n},y_n)$ are exchangeable, CP computes a bound $b_\alpha:\mathbb{R}^{n-m}\to\mathbb{R}$ such that
    $\mathbb{P}\big(\ncf (x,y)\le b_\alpha(s_{m+1},\dots, s_{n})\big)\ge 1-\alpha,$
where $s_i = \ncf (x_i,y_i)$. From this bound over nonconformity scores, we can build a \emph{prediction region} over target values, $\Gamma^{\alpha}(x;\mathcal{I}_C)$, by including all targets $y'$ whose nonconformity scores are likely to follow the distribution of nonconformity scores for the true targets, i.e., have a nonconformity score lower than $b_\alpha(s_{m+1},\dots, s_{n})$:
\begin{equation}\label{eq:cp_predr}
\mathclap{\Gamma^{\alpha}(x; \mathcal{I}_C) := \left\{y'\in \mathcal{Y}: \ncf (x, y')\le b_\alpha(s_{m+1},\dots, s_n)\right\}.}
\end{equation}
This prediction region is guaranteed to contain the true (unknown) value $y$ with confidence $1-\alpha$:
\begin{equation}\label{eq:cp_guar}
\mathclap{    \mathbb{P}\big(y\in \Gamma^\alpha(x;\mathcal{I}_C)\big)\ge 1-\alpha. }
\end{equation}
We stress that the above guarantees are marginal, meaning averaged over the test and calibration data. They hold when $(x,y)$ is exchangeable w.r.t. calibration data $\mathcal{I}_C$, i.e., when the joint probability of $(x_{m+1},y_{m+1}), \dots, (x_{n},y_{n}), (x, y)$ is invariant to permutations.
A natural choice to derive the upper bound $b_\alpha$
from the calibration nonconformity scores $s_{m+1},\dots s_{n}$ is the quantile function, i.e., $b_\alpha (s_{m+1},\ldots, s_{n}) := \mathcal{Q}_{1-\alpha}(\mathcal{F}_C)$. More precisely, given $\alpha\in (0,1)$, $\mathcal{Q}_{1-\alpha}(\mathcal{F}_C)$ denotes the $(1-\alpha)$-th quantile over the empirical distribution $\mathcal{F}_C = \tfrac{1}{n-m+1}\big(\sum_{(x_i,y_i)\in \mathcal{I}_C} \delta_{s_i}+\delta_\infty\big)$, with $\delta_{s_i}$ being the Dirac distribution centered at $s_i=\ncf (x_i, y_i)$ and $\infty$ is added as a correction to
obtain finite sample guarantees. For a generic real-valued random variable $T$, the quantile function $\mathcal{Q}$ is formally defined as $\mathcal{Q}_{1-\alpha}(T) = \inf \{t\in\mathbb{R}\mid \mathbb{P}(T\le t)\ge 1-\alpha\}$.
A common choice as NCF for \emph{regression} problems is the Euclidean distance, i.e.,  
    $\ncf (x_i, y_i)=\| y_i-\hat{f}\left(x_i\right)\|_2.$
In \emph{quantile regression} problems~\cite{romanoCQR}, where the model 
outputs a quantile range
, i.e., $\hat{f}(x)= [\hat{q}_{l}(x),\hat{q}_{u}(x)]\subseteq\mathbb{R}$, a common NCF is 
$\ncf (x_i, y_i)= \max \{\hat{q}_{l}(x_i)-y_i, y_i-\hat{q}_{u}(x_i)\}.$ 
For these particular NCF choices, 
the predictive interval  
is defined as:
\begin{equation}
    \label{cpInterval}
   \mathclap{ \Gamma^{\alpha}\left(x;\mathcal{I}_C\right) = \left[ \hat{q}_{l}(x) - \mathcal{Q}_{1-\alpha}(\mathcal{F}_C), \hat{q}_{u}(x) + \mathcal{Q}_{1-\alpha}(\mathcal{F}_C)\right],}
\end{equation}
where 
$\hat{q}_{l}(\cdot)=\hat{q}_{u}(\cdot)=\hat{f}(\cdot)$ in case of deterministic predictors. Intuitively, $\Gamma^{\alpha}\left(x;\mathcal{I}_C\right)$ enlarges or shrinks the quantile range by adding a value of $\mathcal{Q}_{1-\alpha}(\mathcal{F}_C)$ to both quantiles to correct for miscoverage.
The guarantees in~\eqref{eq:cp_guar} hold regardless of the underlying data distribution, the choice of predictive model $\hat{f}$, or the accuracy of the chosen model \cite{introToCP}.  

\subsection{Copulas and the Mahalanobis Distance}
\label{subsec:copulas}
Let $X = (X_1, \dots, X_d)$ be a multivariate random vector, with $F(x) = P(X_1\leq x_1,\dots, X_d\leq x_d)$ its joint Cumulative Distribution Function (CDF). Additionally, let $F_1,\dots, F_d$ be the marginal CDFs of $X_1, \dots, X_d$. By applying the probability integral transform to each element of $X$, we obtain the random vector $(U_1,\dots, U_d)=(F_1(X_1),\dots,F_d(X_d))$ whose marginals follow a uniform distribution on $[0,1]$. With his work in 1959 \cite{sklar1959fonctions}, Sklar investigated the relation between joint CDFs and their marginals through copulas.
\begin{definition}
A copula is a function $C:[0,1]^d\to[0,1]$ which satisfies the following properties:
\begin{enumerate}
    \item For every $u = (u_1, \dots u_d)\in[0,1]^d$, if there exists $j\in\{1,\dots,d\}$ such that $u_j = 0$, then $C(u)=0$.
    \item For any $j\in\{1,\dots,d\}$ and for any $u_j\in[0,1]$, if $u = (1, \dots, 1, u_j,1, \dots, 1)$, then $C(u)=u_j$.
    \item $C$ is d-increasing: for any $a = (a_1, \dots, a_d)$ and $b=(b_1,\dots,b_d)$ in $[0,1]^d$ such that $a_j\leq b_j$ for all $j\in\{1,\dots,d\}$, the C-volume $\Delta_{(a, b]}C$ of the hyper-rectangle defined between a and b is non negative \cite{nelsen2006introduction}.
\end{enumerate}
\end{definition}
\fc{Sklar's Theorem (Appendix~\ref{app:sklar}) establishes that copulas link marginal distributions to their joint distribution, enabling flexible modeling of multivariate dependencies independent of individual marginals. Since marginals are often easier to estimate, copulas alone suffice to model the joint distribution, allowing practitioners to choose from diverse copula families to capture varying dependency structures.}
We here introduce three different types of copula, defined for any $(u_1,\dots, u_d)\in[0,1]^d$.
\begin{itemize}
    \item \textbf{Independence Copula} represents the dependence structure of independent random variables and corresponds to the scenario in which the joint distribution is simply the product of its marginals:
    $C^{I}(u_1, \dots, u_d) = \prod_{i=1}^{d}u_i$.
    \item \textbf{Gaussian Copula} models the full joint distribution as a multivariate normal entirely defined by the correlation matrix $\Sigma\in[-1,1]^{d\times d}$: 
    \begin{equation}\label{gaussianCopula}
    \mathclap{C^{Gauss}_{\Sigma}(u_1, \dots,u_d) = \Phi_{\Sigma}\left(\Phi^{-1}(u_1),\dots,\Phi^{-1}(u_d)\right),}
\end{equation}
 where $\Phi^{-1}(\cdot)$ is the inverse of a univariate standard normal CDF and $\Phi_{\Sigma}(\cdot)$ is the joint CDF of a multivariate normal distribution with mean vector zero and covariance matrix equal to the correlation matrix $\Sigma$. 
    
    \item \textbf{Student's t Copula} models the full joint distribution as a joint CDF of the multivariate Student's t CDF with a zero mean vector, $\nu$ degrees of freedom, and a scale or covariance matrix equal to the correlation matrix $\Sigma$:
    \begin{equation}
    \label{studentCopula}
    \mathclap{C^{Stud}_{\Sigma, \nu}(u_1, \dots, u_d)=t_{\Sigma, \nu}\left(t^{-1}_{\nu}(u_1),\dots ,t^{-1}_{\nu}(u_d)\right),}
\end{equation}
where $t^{-1}_{\nu}(\cdot)$ is the inverse of a univariate Student's t CDF with $\nu$ degrees of freedom.
\end{itemize}
\begin{remark}
The Gaussian copula assumes symmetric dependencies but lacks tail dependence, limiting its ability to model joint extreme events. In contrast, Student’s t copula captures symmetric tail dependence, controlled by its degrees of freedom ($\nu$). Lower $\nu$ strengthens tail dependence, while $\nu \to \infty$ recovers the Gaussian copula with vanishing tail dependence.
\end{remark}

\fc{The \emph{Mahalanobis distance} measures how far an observation deviates from a known multivariate distribution. For a point 
$x\in\mathbb{R}^d$ and a distribution with mean $\mu$ and a covariance matrix $\Sigma$, the Mahalanobis distance $\mahaDist{x}:\mathbb{R}^d\to\mathbb{R}$ is defined as $\mahaDist{x} = \sqrt{\left(x - \mu\right)^T\cdot\Sigma^{-1}\cdot\left(x-\mu\right)}.$
By using the inverse covariance matrix, it accounts for variable correlations, making it effective for outlier detection. If the distribution is multivariate normal, $\mahaDistSquare{x}$ follows a 
$\chi^2$-distribution with $d$ degrees of freedom \cite{ghorbani2019mahalanobisGaussian}; for a multivariate Student's t-distribution, it follows a scaled 
$\mathcal{F}$-distribution with $d,\nu$ degrees of freedom \cite{roth2012multivariate}.}

\section{Methodology}
\label{sec:methodology}
\begin{figure}[!b]
    \centering
    \vspace{-.5cm}
    \includegraphics[width=0.95\linewidth]{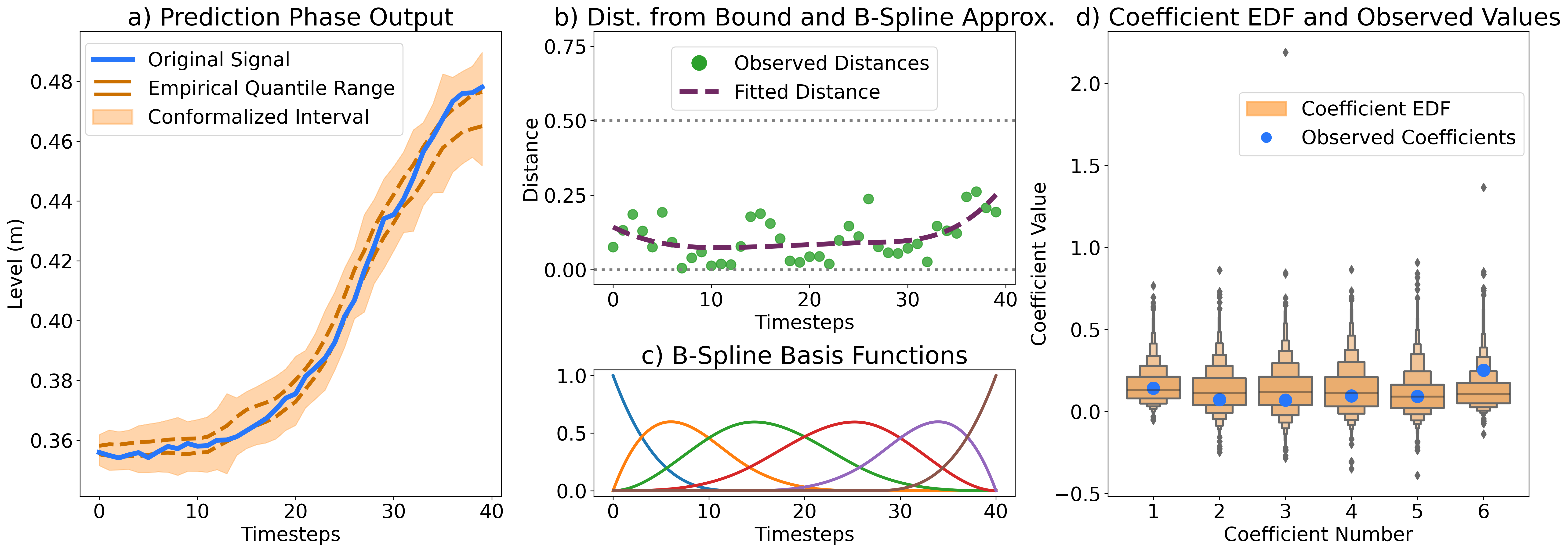}
    \vspace{-.25cm}
    \caption{\textbf{(a)}: Ground truth (blue line), outputs of the quantile regressor (orange dashed line) and conformalized predictive region (orange shading); \textbf{(b,c)}: observed distances between predicted and observed values (green dots) vs. fitted using B-splines (purple dashed line); \textbf{(d)}: Empirical distribution of spline coefficients over an held out dataset (orange), observed coefficients for a test point (blue dots).}\label{fig:cocai-phases}
\end{figure}

The proposed methodology (CoCAI) presents a novel and flexible framework which allows practitioners to obtain accurate predictive 
regions
which offer statistically sound coverage guarantees for any target segment in a MTS. Starting from these 
regions, CoCAI also provides a statistically meaningful indicator allowing for the identification of out-of-distribution, and thus potentially anomalous, observations. 
\np{CoCAI's main advantage lies in its flexible, three-phase pipeline, which can be tailored to better suit the specifics of the problem at hand.}
In the first part of the procedure, the \emph{predictive phase}, an imputation model is used to obtain predictions for the target segments of the MTS. 
\np{For this, we adopt the modified CSDI model from \cite{pearson2025udm} to perform quantile regression on target data.} In the subsequent \emph{conformalization phase}, the resulting empirical quantile range is adjusted through a CP procedure specifically tailored for time-series, yielding a predictive 
region which offers statistically rigorous coverage guarantees. Fig.~\ref{fig:cocai-phases}-(a) \np{shows the ground truth (blue), the quantile regressor outputs (dashed orange) and the conformal predictive region (orange shaded), which provides coverage guarantees. }
In the final \emph{anomaly detection phase}, the distances between predicted and observed values are computed using a metric of choice (green dots in Fig.~\ref{fig:cocai-phases}-(b)).
A dimensionality reduction technique which preserves temporal dependencies is then used on the resulting series, which, combined with copula models, provides a statistically sound indicator on how out-of-distribution, and thus potentially anomalous, a datapoint is, as depicted in Fig.~\ref{fig:cocai-phases}-(b,c,d). It is important to note that the predictive phase is independent from the rest of the procedure, meaning that any kind of imputation model can be used as long as a suitable distance metric is used in the anomaly detection phase. Additionally, both the conformalization and anomaly detection phases require a one-time offline calibration step that does not require repetition during deployment. In the following, each component of the proposed pipeline will be described in detail.
\subsection{Prediction Phase}
\label{predictionPhase}
\begin{figure}[!b]
     \centering
     \vspace{-.75cm}
     \includegraphics[width=0.95\textwidth]{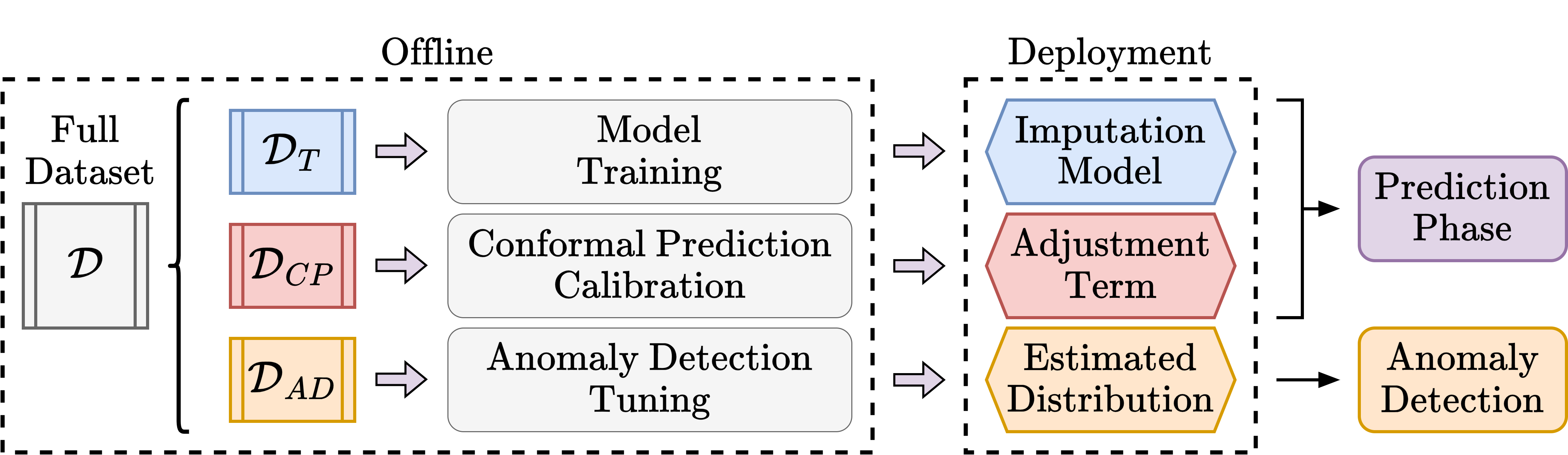}
     \vspace{-.25cm}
     \caption{Scheme of the CoCAI pipeline highlighting the offline and deployment steps}
     \label{fig:data_split}
 \end{figure}

Given a dataset of $n$ elements $\fulldataset = \{\fullseries_{1:T}^{i}\}_{i=1}^{n}$, where each $\fullseries_{1:T}^{i}\in \mathbb{R}^{T\times d}$ is a 
 multivariate time-series with $d$ channels and $T$ time steps, let $\mathcal{J}\subseteq \{1, \dots, d\}$ be a non-empty index set such that $|\mathcal{J}|=d'$ and $d'\leq d$. Given $t\in \mathbb{N}$, where $1\leq t\leq T$, we can define a prediction target as $\target^{i}_{1:t,  \mathcal{J}}= \{\fullseries_{\tau, j}^i:\tau\in\{T-t,\dots,T\}, j \in \mathcal{J}\}$ with $\target^{i}_{1:t, \mathcal{J}}\in \mathbb{R}^{t\times d'}$ and $\fullseries_{\tau, j}^i$ being the value of the $j$-th channel of the $i$-th series at time step $\tau$. For simplicity, in the following we will refer to the $i$-th complete MTS $\fullseries_{1:T}^{i}$ as $\fullseries^i$, while its prediction target $\target^{i}_{1:t, \mathcal{J}}$ will be referred to as $\target^i$. Lastly we denote all of the elements of $\fullseries^i$ which are not included in $\target^i$ as $\cond^i$. In this way we can express the full MTS as $\fullseries^i = \cond^i \cup \target^i$, with $\cond^i \cap \target^i = \emptyset$, and allowing for a clear separation between the conditioning information $\cond^i$ from the prediction targets $\target^i$. As 
 shown in Fig.~\ref{fig:data_split}, the complete dataset $\fulldataset$ is split into three subsets: a training set $\datasetTrain$ used to train an imputation model, and two calibration sets $\datasetCP$ and $\datasetAD$ used respectively in the conformal prediction 
 and the anomaly detection phase.
 \newline Following \cite{pearson2025udm}, the CSDI framework is used to obtain plausible predictions for our multivariate time-series data, with the objective of obtaining imputations for the target data $\target^i$, conditioned on the observed values $\cond^i$. At training time CSDI learns an estimate for $p_\theta\left(\target^i\mid\cond^i\right)$, the conditional probability distribution of the target values of the time series conditioned on the observed ones, from $\datasetTrain$. This is achieved using a self-supervised procedure where various portions of the MTS are masked or obscured and the model attempts to reconstruct them given the remaining observations. The choice of which parts to mask plays an essential role in defining what data patterns the model will be able to reconstruct in the inference phase and should ideally match the objective of the target application and the expected missing data patterns. At inference time, the model generates multiple plausible imputations for the target data by repeatedly sampling from the estimated conditional distribution, using different random noise initializations. 
 Each sample produces a  distinct realization of the target value $\target^i$, naturally capturing the uncertainty in the imputation process. 
 From these imputations, we can derive either point estimates (using aggregate statistics like the mean or median) or uncertainty estimates (through empirical quantile ranges). While our methodology primarily focuses on quantile-based approaches due to their superior ability to capture imputation uncertainty, the proposed pipeline remains equally applicable to point estimates with minimal modifications. 
More precisely, for each target channel $j\in\mathcal{J}$ and each time step $\tau \in\{T-t,\dots, T\}$, the $\epsilon$-th quantile function, $\hat{q}^{\tau,j}_{\epsilon}(x^i)$, is computed as the $\epsilon$-th empirical quantile extracted from the learned probabilistic output $p_\theta(y^i|x^i)$ at time $\tau$ and channel $j$.
 Given a desired confidence level $1-\alpha$, we can denote an empirical quantile range as $\left[\quantileLow{\cond^i}, \quantileHigh{\cond^i}\right]$ obtained by extracting, from the set of sampled imputations, the quantiles of order $\alpha/2$ and $1-\alpha/2$ for each target channel and time step independently. In practice, $\quantileLow{\cond^i}$ and $\quantileHigh{\cond^i}$ are two $(t\times d')$-dimensional objects, i.e., $\hat{q}_*({\cond^i}) = \{\hat{q}_*^{\tau,j}({\cond^i})\}_{\tau,j}$ for $*\in\{l,u\}$.
 The predictive region constructed in this way is a sequence of length $t$ of $d'$-dimensional hyper-rectangles and, in general, does not provide the coverage guarantee of containing the ground truth with the desired confidence level. More formally, the extracted quantile range does not satisfy $\mathbb{P}\left(\target^i\in\left[\quantileLow{\cond^i}, \quantileHigh{\cond^i}\right]\right)\geq 1-\alpha$, where the condition $\target^i\in\left[\quantileLow{\cond^i}, \quantileHigh{\cond^i}\right]$ is true only if, for every $\tau\in\{T-t,\dots, T\}$, the true target $\target^i_\tau$ falls inside the hyper-rectangle defined by the quantile range at time $\tau$. To address this limitation, we use CopulaCPTS \cite{sun2023copulaCPTS}, a CP copula based approach tailored for multivariate and multi time-step time-series imputation tasks. CopulaCPTS leverages copula models to provide confidence intervals which are as small as possible while still satisfying the formal coverage guarantee. We adapt CopulaCPTS to operate on the probabilistic output of CSDI, constructing intervals starting from empirical quantile ranges. To this end we utilize the non-conformity score proposed in \cite{romanoCQR}:
    $\ncf (x^i, y^i)= \max \{\hat{q}_{l}(x^i)-y^i, y^i-\hat{q}_{u}(x^i)\}.$
\begin{figure}[!b]
    
    \centering
    \vspace{-.5cm}\includegraphics[width=0.95\textwidth]{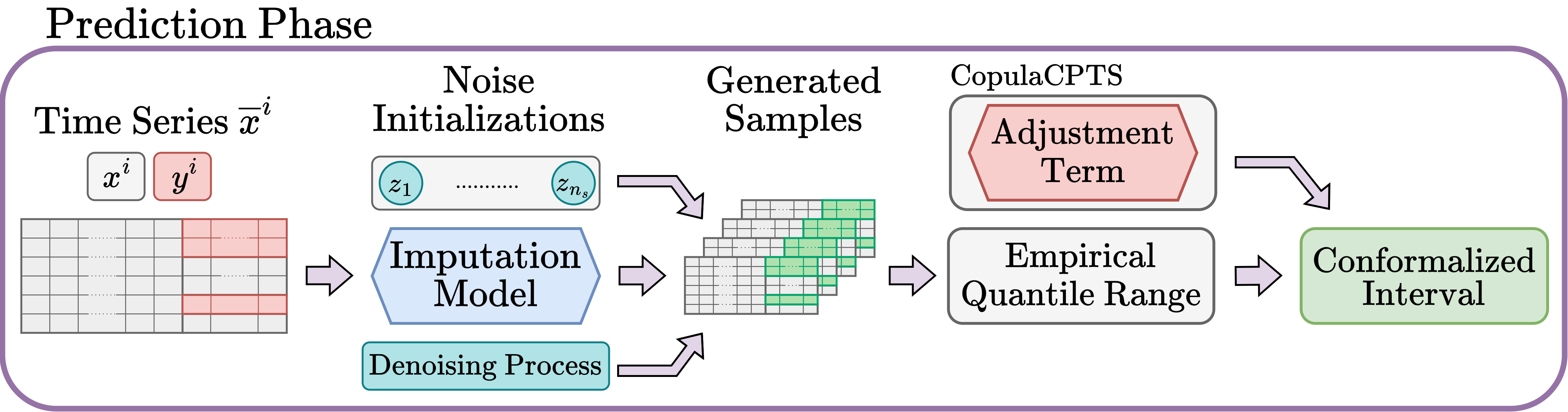}
    \vspace{-.25cm}
    \caption{Pipeline of the full prediction phase}
    \label{fig:predictionPhase}\vspace{-.5cm}
\end{figure}
Additionally, we replaced the gradient-descent optimization of~\cite{sun2023copulaCPTS}, which in our experiments often led to unstable results, with a bounded optimization algorithm. The latter explicitly encodes efficiency as the objective to maximize, while 
imposing validity constraints. Fig.~\ref{fig:boundedCopulaCPTS} in Appendix~\ref{app:copulaCPTS} shows a qualitative evaluation demonstrating how our solution outperforms the gradient-based one proposed in~\cite{sun2023copulaCPTS}. 
By applying this adapted version of CopulaCPTS in the conformalization process on $\datasetCP$, we obtain an adjustment term (assuming different values at each time step) that, when added to and subtracted from the upper and lower bounds of the original quantile range, provides the predictive interval $\left[\CquantileLow{\cond^i},\CquantileHigh{\cond^i}\right]$. This interval is guaranteed to contain the ground truth values $\target^i$ over all time steps with the desired confidence level $1-\alpha$, while requiring the smallest possible adjustments to the original bounds.
\subsection{Anomaly Detection Phase}
\label{sub:adPhase}
\noindent\textbf{Offline Tuning.}
Let $\datasetAD=\{\fullseries^i\}_{i=1}^{m}$ be a held-out dataset of MTS 
that do not present anomalies in their target data $\{\target^i\}_{i=1}^m$. Following the pipeline described in Sect.~\ref{predictionPhase}, we obtain a conformalized predictive region $\left[\CquantileLow{\cond^i}, \CquantileHigh{\cond^i}\right]$ for each observation of $\datasetAD$. For each observation $\fullseries^i$, we 
compute the distance between the ground truth $\target^i$ and the closest bound of the corresponding interval for each target channel $j$ and target time-step $\tau$ using $\delta^i_{\tau, j}$ defined as:
\begin{equation}
    \label{distanceFunction}
   \mathclap{ \deltaObs{i}{\tau}{j} = -\cfrac{d^i_{\tau, j}}{w^i_{\tau, j}} +0.5,\text{ with }
    d^i_{\tau, j} = \max\left(\CquantileLow{\cond^i}_{\tau, j} - \target^i_{\tau, j}, \target^i_{\tau, j} - \CquantileHigh{\cond^i}_{\tau, j}\right),} 
\end{equation}
and where $w^i_{\tau, j} = \CquantileHigh{\cond^i}_{\tau, j} - \CquantileLow{\cond^i}_{\tau, j} > 0$ is the width of the conformalized interval at target time-step $\tau$ and target channel $j$ for observation $i$. 
This formulation ensures that $\deltaObs{i}{\tau}{j}$ is non-negative, taking values close to $0$ when $\target^i_{\tau, j}$ is near the midpoint of the interval and reaching a value of $0.5$ as $\target^i_{\tau, j}$ gets closer to either bound. Values greater than $0.5$ indicate that the observation is outside the interval. For each observation $i$ and target channel $j$, we can define a time series $\deltaSeq{i}{j}=\left(\deltaObs{i}{1}{j},\dots, \deltaObs{i}{t}{j}\right)$ which represents the relative position of the ground truth with respect to its predictive interval over the full target window. As the length of the target window increases, so does the value of $t$, leading to higher computational costs. 
In order to reduce the dimensionality of $\deltaSeq{i}{j}$, 
we choose a set of B-splines consisting of $K$ cubic functions $\{\phi_k(\cdot)\}_{k=1}^K$ defined over the full target window with uniformly distributed knots. These basis functions are kept constant for all observations and target channels. By using $\{\phi_k(\cdot)\}_{k=1}^K$, we can obtain a smooth curve that approximates each term of the sequence $\deltaSeq{i}{j}$ as:
\begin{equation}
    \label{bspline}
    \mathclap{\deltaFit{i}{\tau}{j} = \sum_{k=1}^{K}\betaObs{i}{k}{j}\cdot\phi_k\left(\tau\right).}
\end{equation}
The resulting curve maintains the structure of the original sequence $\deltaSeq{i}{j}$ and is entirely defined by the coefficient array $\betaSeq{i}{j}=\left(\betaObs{i}{1}{j},\dots,\betaObs{i}{K}{j}\right)$ of length $K$, with $K\leq t$. Given that by construction each basis function $\phi_k\left(\cdot\right)$ is non-zero only over a specific portion of the time domain, the corresponding coefficient $\betaObs{i}{k}{j}$ captures the behavior of $\deltaSeq{i}{j}$ in that particular segment, preserving existing temporal dependencies. To determine the optimal number of basis functions $K$, we compute the residual sum of squares between the observed value $\deltaObs{i}{\tau}{j}$  in $\datasetAD$ and its estimate $\deltaFit{i}{\tau}{j}$ obtained with a range of different values of $K$. By representing these values with an elbow plot, we can identify the value of $K$ which provides an optimal balance between reconstruction accuracy and number of parameters.\newline
\indent By performing this procedure for each observation of $\datasetAD$, we obtain $m$ $K$-dimensional sequences $\betaSeq{i}{j}$ which approximate the original $t$-dimensional $\deltaSeq{i}{j}$. It is crucial to remark that $\datasetAD$ exclusively consists of series which have been deemed to be anomaly-free. Consequently, the derived coefficient sequences $\betaSeq{i}{j}$ provide a low-dimensional representation of normal system behavior, capturing the patterns which can be observed in ordinary, non-anomalous conditions. Provided that the tuning dataset $\datasetAD$ contains a sufficiently large number of representative observations, it is possible to accurately approximate the distribution of each coefficient $\betaObs{}{k}{j}$ using its empirical distribution function (EDF) $\estEDF{k}{j}{\cdot}$. Although these EDFs accurately capture the behavior of each coefficient $\betaObs{}{k}{j}$ independently, they fail to account for correlations or dependencies that may exist with other elements of the full coefficient vector. To address this limitation, we use the estimated EDFs $\estEDF{k}{j}{\cdot}$ as a starting point for fitting Gaussian and Student's t copula models, providing an approximation of the joint distribution of the coefficients and preserving their dependency structure. Each observed coefficient $\betaObs{i}{k}{j}$ is transformed using its corresponding EDF, yielding values $\uObs{i}{k}{j}= \estSEDF{k}{j}{\betaObs{i}{k}{j}}$ which are, by construction, uniformly distributed on the interval $\left[0,1\right]$. By repeating this procedure for each observation $i$ and each coefficient index $k$, we obtain a set of $m$ sequences $\uSeq{i}{j} = (\uObs{i}{1}{j}, \dots,\uObs{i}{K}{j})$ which offer a suitable input for modeling both Gaussian and Student's t copulas. In the case of the Gaussian copula, the sequences $\uSeq{i}{j}$ are further transformed by applying the inverse of a univariate standard normal cumulative density function (CDF) to each of their elements, obtaing $\zSeq{i}{j}=(\zObs{i}{1}{j}, \dots,\zObs{i}{K}{j})$, where $\zObs{i}{k}{j}=\Phi^{-1}(\uObs{i}{k}{j})$. The normalized values $\zSeq{i}{j}$ are used to estimate the empirical covariance matrix $\SigmaFit\in \mathbb{R}^{K\times K}$, obtained by computing the Pearson correlation across all $m$ observations for each pair of coefficients. Recalling Section \ref{subsec:copulas} and Equation \ref{gaussianCopula}, we obtain a Gaussian copula which is uniquely determined by the correlation matrix $\SigmaFit$. This copula provides an estimate for the joint distribution of the coefficient set $\betaSeq{}{j}$ in the form of a multivariate normal distribution with mean vector zero and covariance matrix $\SigmaFit$. A similar procedure can also be employed for the estimation of Student's t copula, only requiring minor adjustments, namely the usage of the inverse  Student's t CDF with $\nu$ degrees of freedom when computing $\zObs{i}{k}{j} = t^{-1}_{\nu}(\uObs{i}{k}{j})$. In addition to the correlation matrix $\SigmaFit$, this kind of copula also requires the estimation of the degrees of freedom parameter $\hat{\nu}$, determined through maximum likelihood estimation. Following Equation \ref{studentCopula}, the estimate for the 
joint distribution of $\betaSeq{}{j}$ takes the form of a multivariate Student's t CDF with a zero mean vector, $\hat{\nu}$ degrees of freedom, and a scale parameter $\SigmaFit$.
\newline\textbf{Deployment.}
 Starting from these estimated joint distribution functions, it is possible to provide an anomaly score for a selected channel $j$ of the target data $\target^*$ of a previously unseen MTS $\fullseries^*$. To ensure the validity of the process, it is essential that the target data $\target^*$ is defined using the same subset of channels $\mathcal{J}$ and target length $t$ as in the previous calibration procedure. Following the predictive procedure described in Fig.~\ref{fig:predictionPhase}, we obtain a conformalized predictive region $\left[\CquantileLow{\cond^*}, \CquantileHigh{\cond^*}\right]$, which is then used to compute the series of distances to the target data $\target^*$, as defined in Equation \ref{distanceFunction}. 
 \begin{figure}[!t]
\vspace{-.5cm}
    \centering
    \includegraphics[width=\textwidth]{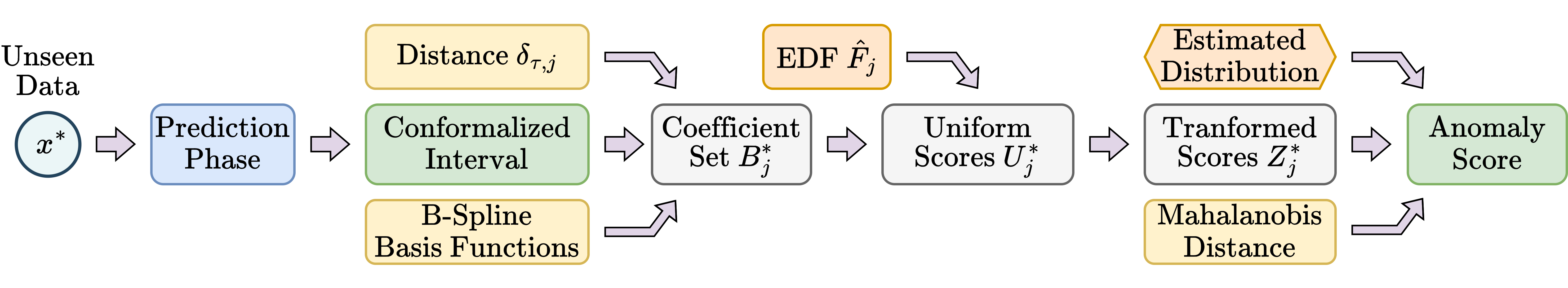}
    \vspace{-.5cm}
    \caption{Schema representing the full deployment pipeline of the proposed methodology}
    \vspace{-.5cm}
    \label{fig:deployment}
\end{figure}
 Using the B-spline basis functions described in~\eqref{bspline}, we obtain a $K$-dimensional representation of our distance series in the form of the coefficient series $\betaSeq{*}{j}$. Each of its elements is transformed into a uniform value $\uObs{*}{k}{j}=\estSEDF{k}{j}{\betaObs{*}{k}{j}}$ using the EDFs estimated in the tuning phase. These uniform values are then further transformed, yielding a score array $\zSeq{*}{j}$, where $\zObs{*}{k}{j}=\Phi^{-1}(\uObs{*}{k}{j})$ when a Gaussian copula is selected and $\zObs{*}{k}{j}=t^{-1}_{\hat{\nu}}(\uObs{*}{k}{j})$ when a Student's t copula is used. We finally use these transformed values to compute an anomaly score by means of the Mahalanobis distance, a distance metric which quantifies the deviation of an observation from an existing and known distribution \cite{mclachlan1999mahalanobis}. 
 \fc{Following the definition in Sect.~\ref{subsec:copulas}, since our copula-estimated multivariate normal and Student’s t distributions have zero mean and covariance matrix $\SigmaFit$, the squared Mahalanobis distance for an observation $\zSeq{*}{j}$ is $\mahaDistSquare{\zSeq{*}{j}} = \left(\zSeq{*}{j}\right)^T\cdot\SigmaFit^{-1}\cdot\left(\zSeq{*}{j}\right)$. For a $K$-dimensional multivariate normal, $\mahaDistSquare{\zSeq{*}{j}} \sim \chi^{2}_{K}$; for a Student’s t with $\nu$ degrees of freedom, $1/\nu\cdot \mahaDistSquare{\zSeq{*}{j}}\sim\mathcal{F}_{K, \nu}$. These distributions allow us to set statistical thresholds for out-of-distribution detection.} \np{The complete deployment pipeline is illustrated in Fig.~\ref{fig:deployment}, detailing each step of the proposed methodology.}

\section{Experiments}
\label{sec:experiments}
\lb{To demonstrate the practical applicability of our approach, we now present experimental results using real-world data. Idrostudi srl has provided us with two real-world datasets which contain operational data from urban infrastructure monitoring campaigns. We have deliberately focused our experimental analysis on this data as it presents greater complexity and reflects the challenges encountered in real-world scenarios.}
\begin{figure}[!b]
\vspace{-.5cm}
    \centering
    \includegraphics[width=0.8\linewidth]{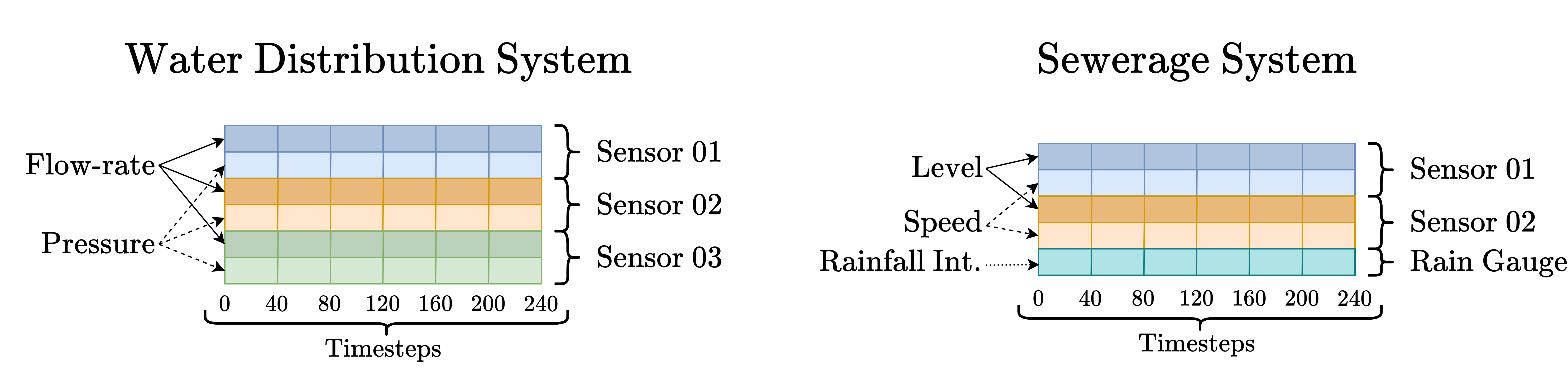}
    \vspace{-.5cm}
    \caption{Graphical representation of the multivariate time-series used in the experimental section for water distribution (left) and sewerage (right) systems. \vspace{-.5cm} }
    \label{fig:WDS_SEW_Setup}
\end{figure}
\np{The first dataset focuses on a Water Distribution System (WDS), while the second one is comprised of time-series derived from sewerage systems, \fz{also referred to as wastewater networks}. Both systems  are essential for local communities, but rely on aging networks which are difficult to maintain. Urban population growth, combined with more frequent extreme weather events due to climate change and widespread soil sealing are additional stressors for sewerage systems, potentially causing overflows with significant environmental and health risks. For WDSs, undetected leakages and pipe breaks can cause water loss, contamination and soil erosion, posing structural risks to nearby infrastructure. To monitor these systems, Idrostudi srl teams conduct campaigns which leverage strategically placed sensors, recording physical parameters every six minutes and providing MTS which accurately reflect the networks behavior.}
\subsection{Case Studies}
\paragraph{Water Distribution Systems.}
For the WDS section of our experimental analysis, we focus on monitoring campaigns performed in Northern Italy, using data recorded by groups of three sensors which are \fz{placed in} topologically connected \fz{locations} within the same network. For each sensor, the physical parameters of interest are water \emph{pressure} and \emph{flow-rate} which were both used in our analysis, yielding a MTS with a total of six channels (as depicted in Fig.~\ref{fig:WDS_SEW_Setup}--left). This grouping strategy was selected to exploit both the correlations existing between different physical parameters measured by the same sensor and the spatial and temporal correlations between the recordings of topologically connected sensors. 

\paragraph{Sewerage Systems.} For the sewerage system  
we adopt a similar grouping strategy, 
pairing sensors which are \fz{placed in} topologically connected \fz{sites} within the same network into groups of two. In this scenario, the physical parameters of interest are the sewage's \emph{level} and \emph{speed}. Moreover, given that wastewater networks are also responsible for collecting stormwater, we also include rainfall intensity measurements taken from a surface rain gauge placed in proximity to the selected sensors. This setup results in a MTS with a total of five channels  (as depicted in Fig.~\ref{fig:WDS_SEW_Setup}--right), which captures both the physical dynamics of the network and the effect of external environmental factors.
\subsection{Experimental Setup}
For both the WDSs and sewerage systems scenarios, we construct datasets composed by MTS covering a 24-hour period. Given that a measurement is taken every six minutes, this translates into series with a total of 240 time steps (Fig. \ref{fig:WDS_SEW_Setup}). The presented experiments focus on the task of contextual forecasting, a predictive approach that generates forecasts using historical values of target variables while leveraging both historical and known current information from other variables as additional context. Specifically, we perform imputations over the final four hours, or 40 timesteps, of a single target channel at the time. The chosen time window was selected in order to simulate the frequency of data transmissions from the data-loggers connected to the sensors and to emulate a near real-time anomaly detection framework. It is important to note that in the sewerage system case study, the channel containing rainfall intensity measurements only operates as contextual information and is not considered as a possible target channel. For the imputation task, we trained two separate CSDI models, one for WDSs and one for sewerage systems. 
To increase imputation flexibility, during training, the masking procedure is
extended to allow for concurrent imputations on multiple channels and longer target windows. 
\fz{Considering that during precipitation events runoff is drained by wastewater networks, these systems present markedly different hydraulic behaviors in dry and wet weather conditions, with significantly higher recorded \emph{level} and \emph{speed} values during wet periods.} Given these differences, we perform the CP calibration procedure separately for the two scenarios. This ensures that the calibration provides adjustments which are not over-inflated in dry weather and too conservative in wet conditions. This partition also remains in place for the AD tuning procedure of sewerage data in order to ensure consistency within the results.
\subsection{Results}
\label{sub:results}
We present here results derived from applying the CoCAI framework to both test cases. For each scenario, we selected one sensor and separately applied the full procedure on both \emph{level} and \emph{speed} measurements for sewerage data, and \emph{flow-rate} and \emph{pressure} for WDSs. 
For brevity, for the sewerage system, we only present results in dry conditions. 
Hyperparameter selection was performed independently for the four test cases, with a value of $\alpha = 0.1$ always being used in the CP procedure. This value was chosen as it offered a reasonable trade-off between high coverage and small interval size. The other hyperparameters, i.e. the number of basis functions $K$ and the value of the degrees of freedom parameter $\hat{\nu}$, were selected following what was presented in Sect.~\ref{sub:adPhase} and the chosen values are visible in Table \ref{tab:copula}. 
The degree of abnormality for each observation is quantified by two \emph{anomaly scores}
$a_G := 1 -p_G$ and $a_S = 1 -p_S$, where $p_G$ and $p_S$ correspond respectively to the p-values associated with the Mahalanobis distance of the transformed observation from the distribution derived from the test case-specific Gaussian and Student's t copula models. This formulation provides interpretable scores bounded within the interval $\left[0,1\right]$, with values closer to 1 indicating stronger evidence for an anomaly.
\begin{table}[!b]

    \centering
    \vspace{-.5cm}
    \caption{Comparison of the median, average width, average relative width and data coverage percentage between the Empirical Quantile Range (EQR) and conformalized intervals across the considered datasets. The results highlight superior coverage for the conformalized interval, while only requiring minor adjustments.
}\label{tab:intervals}\vspace{-.25cm}
\adjustbox{max width=\linewidth}{
\begin{tabular}{lcc|ccc|ccc|}
\multicolumn{1}{c}{}                 &                           &        & \multicolumn{3}{c|}{\textbf{EQR}}                                                                                                                                                                       & \multicolumn{3}{c|}{\textbf{Conformalized Interval}}                                                                                                                                                     \\ \cline{2-9} 
\multicolumn{1}{c|}{}                & \multicolumn{1}{c|}{Obs}  & Median & \multicolumn{1}{c|}{\begin{tabular}[c]{@{}c@{}}Avg. \\ Width\end{tabular}} & \multicolumn{1}{c|}{\begin{tabular}[c]{@{}c@{}}Avg. Rel.\\ Width (\%)\end{tabular}} & Cov. (\%)                            & \multicolumn{1}{c|}{\begin{tabular}[c]{@{}c@{}}Avg. \\ Width\end{tabular}} & \multicolumn{1}{c|}{\begin{tabular}[c]{@{}c@{}}Avg. Rel.\\ Width (\%)\end{tabular}} & Cov. (\%)                             \\ \hline
\multicolumn{1}{l|}{Sew. - Level}    & \multicolumn{1}{c|}{1295} & 0.452  & \multicolumn{1}{c|}{0.008}                                                 & \multicolumn{1}{c|}{1.92}                                                           & {\color[HTML]{FE0000} \textbf{1.62}} & \multicolumn{1}{c|}{0.026}                                                 & \multicolumn{1}{c|}{6.10}                                                           & {\color[HTML]{009901} \textbf{91.58}} \\
\multicolumn{1}{l|}{Sew. - Speed}    & \multicolumn{1}{c|}{1301} & 0.358  & \multicolumn{1}{c|}{0.024}                                                 & \multicolumn{1}{c|}{6.84}                                                           & {\color[HTML]{FE0000} \textbf{2.77}} & \multicolumn{1}{c|}{0.058}                                                 & \multicolumn{1}{c|}{17.05}                                                          & {\color[HTML]{009901} \textbf{90.70}} \\
\multicolumn{1}{l|}{WDS - Flow-rate} & \multicolumn{1}{c|}{1613} & 53.27  & \multicolumn{1}{c|}{2.82}                                                  & \multicolumn{1}{c|}{5.14}                                                           & {\color[HTML]{FE0000} \textbf{0.25}} & \multicolumn{1}{c|}{7.14}                                                  & \multicolumn{1}{c|}{13.24}                                                          & {\color[HTML]{009901} \textbf{90.45}} \\
\multicolumn{1}{l|}{WDS - Pressure}  & \multicolumn{1}{c|}{1604} & 7.50   & \multicolumn{1}{c|}{0.03}                                                  & \multicolumn{1}{c|}{0.35}                                                           & {\color[HTML]{FE0000} \textbf{2.24}} & \multicolumn{1}{c|}{0.07}                                                  & \multicolumn{1}{c|}{0.92}                                                           & {\color[HTML]{009901} \textbf{89.29}} \\ \hline
\end{tabular}
\vspace{-.5cm}
}
\end{table}

\begin{figure}[!t] 
  \centering
  \vspace{-.25cm}
     \includegraphics[width=\textwidth]{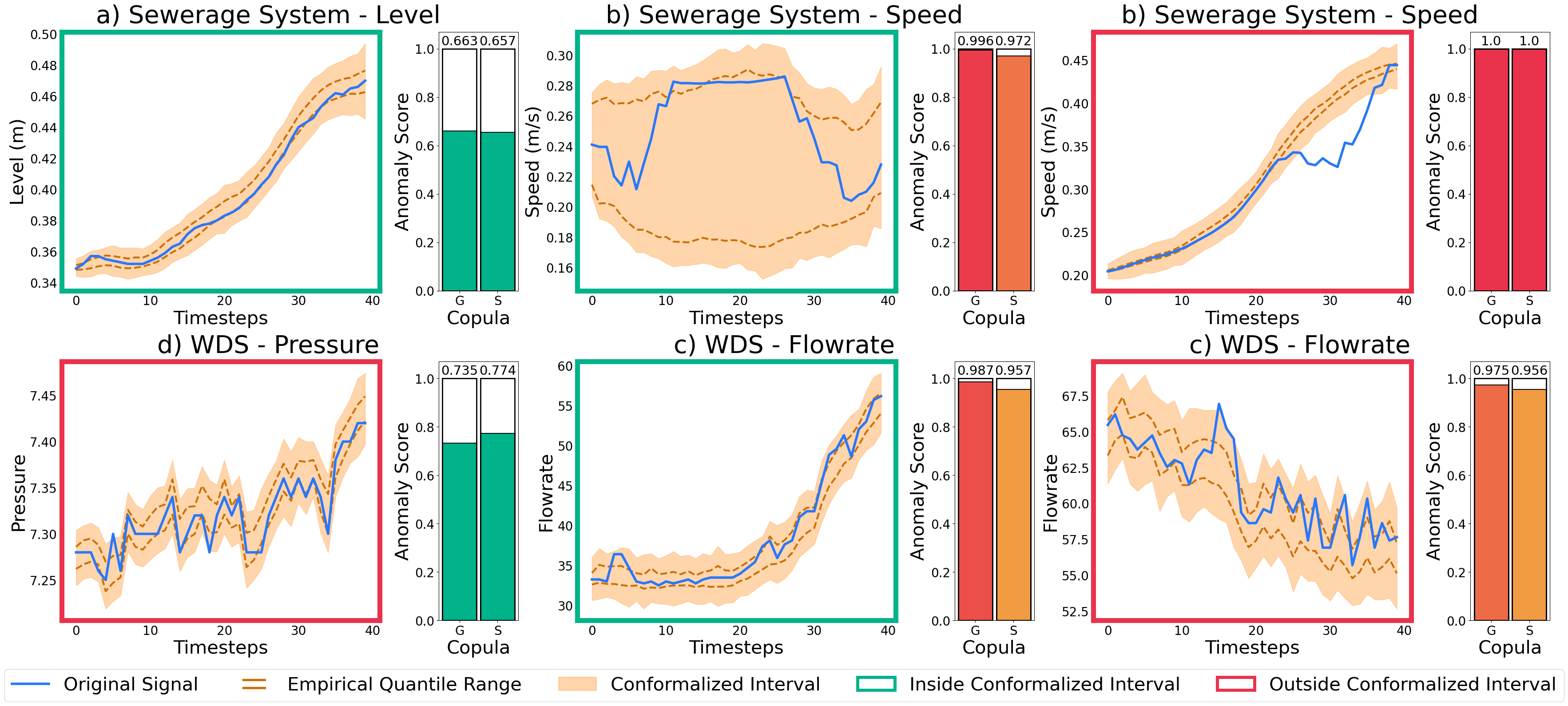}
     \vspace{-.5cm}
     \caption{Prediction and anomaly detection results for a set of time-series derived from sewerage networks \textbf{(a,b,c)} and WDSs \textbf{(d,e,f)}. The left side of each panel shows the conformalized interval (orange shading), EQR (orange dashed line) and the observed target sequence (blue line). Green/red borders indicate if the observed values are fully covered by the conformalized interval. The bar plots in the right side of each panel represent the anomaly scores derived from Gaussian (left) and Student's t (right) copulas.}
     \label{fig:result_plot}\vspace{-.625cm}
 \end{figure}

We evaluate the performance of the CoCAI pipeline from two complementary perspectives. First, we assess the quality of the predictive phase by evaluating the predicted conformalized regions. Table \ref{tab:intervals} shows the coverage rates for both the empirical quantile ranges (EQR) and conformalized regions. These values represent the percentage of test time-series for which the target sequence falls within the respective interval over all timesteps simultaneously. The results show that the conformalized regions not only provide a significant improvement over the EQR, but also achieve the desired coverage rate for all four test cases. While this increase in coverage comes at the cost of marginally wider intervals, their width remains within satisfactory limits when compared to the target value, as can be seen in Fig.~\ref{fig:result_plot}. The figure presents a qualitative evaluation of a selection of examples from all four test cases, where the left plot in each panel shows the ground truth for the target segment (blue line), the EQR (orange dashed line) and conformalized interval (orange shading). Green and red borders indicate if the observed values are fully covered by the conformalized interval or not. Additional plots are shown in Appendix~\ref{app:plots} (Fig.~\ref{fig:result_plot_SEW_00}-~\ref{fig:result_plot_WDS_00}).

Second, we evaluate the anomaly detection component by analyzing which observations are flagged as anomalous by the computed scores. For both $a_G$ and $a_S$ we select a threshold value of $0.9$ and consider as potentially anomalous all series for which at least one of the scores is greater than $0.9$. This value was selected for all four test cases in order to increase comparability between different datasets. Results presented in Table~\ref{tab:copula} and Fig.~\ref{fig:result_plot} show that there is generally strong agreement between the two anomaly scores, with $a_S$ usually proposing a more conservative output. This is to be expected given the heavier tails of the underlying distribution and results in the Student's t copula flagging as anomalous a number of series which is more aligned with what would be expected given the selected threshold. Limited variability within the proportion of flagged series across different test cases can also be observed, likely due to differences in the underlying characteristics of the time-series. In particular, noisier signals such as \emph{speed} in sewerage systems and \emph{flow-rate} for WDS data (Fig.~\ref{fig:result_plot}-(b,f)), are more likely to be flagged as anomalous, especially by the Gaussian score. If necessary, this issue can be mitigated by setting case-specific threshold values as it enables tuning the sensitivity of the anomaly detection process based on the specific context and operational constraints. From Table~\ref{tab:copula} is also possible to note that a target segment which is not fully covered by the conformalized interval is more likely to have a high anomaly score than one which is. However, this is not the only criterion influencing anomaly detection. Fig.~\ref{fig:result_plot} contains a collection of illustrative cases showing different combinations of anomaly scores and interval coverage. In particular we can observe that panels (b,e) show time-series which are fully covered by the conformalized interval but are flagged as potentially anomalous. Conversely, panel (d) shows an example which is not flagged as anomalous even if the observed value falls outside of the interval for at least one time step. This highlights the benefits of using both approaches concurrently, as they offer complementary insights in potentially anomalous data behavior.
\begin{table}[!t]
    \centering
    \vspace{-.25cm}
    \caption{Comparison of flagging rates for both Gaussian and Student's t Copula models while also accounting for conformalized interval coverage rates over the considered datasets. Results for series which are not flagged are also shown.}
\label{tab:copula}\vspace{-.25cm}
\adjustbox{max width=\linewidth}{
\begin{tabular}{lcc|cc|ccc|cc}
\multicolumn{1}{c}{}                 &                                                                       &                                                     & \multicolumn{2}{c|}{\textbf{$\mathbf{a_G > 0.9}$}}                                                                               & \multicolumn{3}{c|}{\textbf{$\mathbf{a_S > 0.9}$}}                                                                                                             & \multicolumn{2}{c}{\textbf{$\mathbf{a_G \leq 0.9 \wedge a_S \leq 0.9}$}}                                                                              \\ \cline{2-10} 
\multicolumn{1}{c|}{}                & \multicolumn{1}{c|}{\begin{tabular}[c]{@{}c@{}}\#\\ obs\end{tabular}} & \begin{tabular}[c]{@{}c@{}}\# \\ basis\end{tabular} & \begin{tabular}[c]{@{}c@{}}Inside\\ Interval (\%)\end{tabular} & \begin{tabular}[c]{@{}c@{}}Outside\\ Interval (\%)\end{tabular} & \multicolumn{1}{c|}{$\nu$} & \begin{tabular}[c]{@{}c@{}}Inside\\ Interval (\%)\end{tabular} & \begin{tabular}[c]{@{}c@{}}Outside\\  Interval (\%)\end{tabular} & \begin{tabular}[c]{@{}c@{}}Inside\\ Interval (\%)\end{tabular} & \multicolumn{1}{c|}{\begin{tabular}[c]{@{}c@{}}Outside\\ Interval (\%)\end{tabular}} \\ \hline
\multicolumn{1}{l|}{Sew. - Level}    & \multicolumn{1}{c|}{1295}                                             & 15                                                  & \begin{tabular}[c]{@{}c@{}}110\\ (8.49)\end{tabular}           & \begin{tabular}[c]{@{}c@{}}73\\ (5.64)\end{tabular}             & \multicolumn{1}{c|}{25}    & \begin{tabular}[c]{@{}c@{}}79\\ (6.10)\end{tabular}            & \begin{tabular}[c]{@{}c@{}}63\\ (4.86)\end{tabular}              & \begin{tabular}[c]{@{}c@{}}1076\\ (83.09)\end{tabular}         & \multicolumn{1}{c|}{\begin{tabular}[c]{@{}c@{}}36\\ (2.78)\end{tabular}}             \\ \hline
\multicolumn{1}{l|}{Sew. - Speed}    & \multicolumn{1}{c|}{1301}                                             & 13                                                  & \begin{tabular}[c]{@{}c@{}}119\\ (9.15)\end{tabular}           & \begin{tabular}[c]{@{}c@{}}89\\ (6.84)\end{tabular}             & \multicolumn{1}{c|}{20}    & \begin{tabular}[c]{@{}c@{}}73\\ (5.61)\end{tabular}            & \begin{tabular}[c]{@{}c@{}}76\\ (5.84)\end{tabular}              & \begin{tabular}[c]{@{}c@{}}1061\\ (81.55)\end{tabular}         & \multicolumn{1}{c|}{\begin{tabular}[c]{@{}c@{}}32\\ (2.46)\end{tabular}}             \\ \hline
\multicolumn{1}{l|}{WDS - Flow-rate} & \multicolumn{1}{c|}{1613}                                             & 15                                                  & \begin{tabular}[c]{@{}c@{}}163\\ (10.11)\end{tabular}          & \begin{tabular}[c]{@{}c@{}}85\\ (5.27)\end{tabular}             & \multicolumn{1}{c|}{11}    & \begin{tabular}[c]{@{}c@{}}91\\ (5.64)\end{tabular}            & \begin{tabular}[c]{@{}c@{}}70\\ (4.34)\end{tabular}              & \begin{tabular}[c]{@{}c@{}}1293\\ (80.16)\end{tabular}         & \multicolumn{1}{c|}{\begin{tabular}[c]{@{}c@{}}68\\ (4.22)\end{tabular}}             \\ \hline
\multicolumn{1}{l|}{WDS - Pressure}  & \multicolumn{1}{c|}{1604}                                             & 15                                                  & \begin{tabular}[c]{@{}c@{}}105\\ (6.55)\end{tabular}           & \begin{tabular}[c]{@{}c@{}}73\\ (5.55)\end{tabular}             & \multicolumn{1}{c|}{23}    & \begin{tabular}[c]{@{}c@{}}72\\ (4.49)\end{tabular}            & \begin{tabular}[c]{@{}c@{}}66\\ (4.11)\end{tabular}              & \begin{tabular}[c]{@{}c@{}}1327\\ (82.73)\end{tabular}         & \multicolumn{1}{c|}{\begin{tabular}[c]{@{}c@{}}99\\ (6.17)\end{tabular}}             \\ \hline
\end{tabular}
}\vspace{-.5cm}
\end{table}

\section{Conclusions}
\label{sec:Conclusions}
In this work we have proposed CoCAI, a novel and flexible framework for multivariate time-series.  CoCAI leverages denoising diffusion probabilistic models combined with conformal prediction techniques to provide accurate predictive regions that offer statistically grounded coverage guarantees. These regions are then used alongside copula-based modeling to derive interpretable anomaly scores, enabling the detection of observations that deviate from the expected pattern. Through practical experiments on real-world operational data derived from sewerage and water distribution systems, we have demonstrated the pipeline's effectiveness in identifying potential anomalies. Furthermore, having the ability to set custom threshold values allows practitioners to tailor the flagging mechanism according to the constraints of the specific application. It is important to note that the current implementation of CoCAI applies the conformal prediction and anomaly detection phases separately for each channel of the multivariate time-series, as it does not yet support a fully multivariate approach for these stages. Given that the underlying predictive model already supports multivariate imputations, extending the full pipeline to operate in a full multivariate setting presents a promising direction for future work. Additionally, we also plan to evaluate the performance of CoCAI on multivariate time-series from other domains, such as medical data, as well as on standardized benchmark datasets. 




\begin{credits}
\subsubsection{\ackname} 
This work has been supported by Idrostudi srl, who is funding the PhD scholarship of Nicholas A. Pearson. The authors also wish to thank Idrostudi srl for providing the operational datasets that enabled the experimental evaluation presented in this work and for their constant support and interest in this project.\newline\newline
This work has been partially supported by the PNRR project iNEST (Interconnected North-Est Innovation Ecosystem) funded by the European Union Next-GenerationEU (Piano Nazionale di Ripresa e Resilienza (PNRR) – Missione 4 Componente 2, Investimento 1.5 – D.D. 1058 23/06/2022, ECS\_00000043).

\subsubsection{\discintname}
The authors have no competing interests to declare that are
relevant to the content of this article. 
\end{credits}

%
%
\bibliographystyle{splncs04}
\bibliography{bibliography}

\newpage

\appendix
\renewcommand{\thefigure}{A\arabic{figure}}
\renewcommand{\thetable}{A\arabic{table}}
\setcounter{figure}{0}
\setcounter{table}{0}

\section{Technical Background}\label{app:bg}

\subsection{Copula Theory}\label{app:sklar}
Sklar's Theorem~\cite{sklar1959fonctions} sets the basis for copula theory as it states the conditions for their existence and uniqueness.

\begin{theorem}
Let $F$ be a d-dimensional CDF,  with marginals $F_1$, ..., $F_d$.\newline
\textbf{(i) Existence:}
There exists a copula, $C$, such that
\begin{equation}\label{sklar:01}
    F(x_1, ..., x_d) = C(F_1(x_1),...,F_d(x_d))
\end{equation} for all $x_j \in [-\infty, \infty]$ and $j\in\{1,\dots,d\}$.\newline \textbf{(ii) Uniqueness:} If $F_j$ is continuous for all $j=1,\dots,d$ then C is unique; otherwise C is uniquely determined only on $Ran(F_1)\times\dots\times Ran(F_d)$ where $Ran(F_j)$ denotes the range of the marginal CDF $F_j$. Conversely, if C is a copula and $F_1, \dots, F_d$ are univariate CDF's, then F defined as in (\ref{sklar:01}) is a multivariate CDF with marginals $F_1,\dots,F_d$.
\end{theorem}

\subsection{Score-based Diffusion Models}\label{app:ddpm}

Denoising Diffusion Probabilistic Models (DDPMs) \cite{hoDDPM} are a type of deep generative models that have been shown to produce high quality samples in a variety of fields. 
DDPMs are trained in a two step procedure to learn an approximation of a true data distribution. In the initial forward process, data is gradually and iteratively corrupted by adding small amounts of Gaussian noise via a Markov chain. In the complementary backwards process a neural network is trained to gradually estimate and remove the noise at each step, learning an iterative procedure that allows the generation of clean samples starting from pure random noise. 

More formally, let $x_0$ be data belonging to the unknown target data distribution $q(x_0)$, and let $p_\theta(x_0)$ be a model distribution that aims to approximate $q(x_0)$. The \emph{forward} process is described by a Markov chain over $\mathcal{K}$ steps of form:
\begin{equation}
    \label{fullForwardProcess}
    q\left(x_{1:\mathcal{K}}|x_0\right) = \prod_{k=1}^{\mathcal{K}}q\left(x_k|x_{k-1}\right),
\end{equation}
which progressively corrupts the data with Gaussian noise. At each step:
\begin{equation}
    \label{transitionForwardProcess}
    q\left(x_k|x_{k-1}\right) = \mathcal{N}\left(\sqrt{1-\beta_k}x_{k-1}, \beta_k I\right),
\end{equation}
where $\beta_k$ is a small positive constant which represents the rate of noise injection at step $k$ and $x_k$ is the corrupted version of the original data $x_0$ at step $k$ of the Markov chain. As $k$ approaches $\mathcal{K}$, the distribution of $x_k$ converges to a standard Gaussian distribution $\mathcal{N}\left(0, I\right)$. Furthermore, it is possible to define $q\left(x_k\mid x_0\right)$ in a closed form as $q\left(x_k\mid x_0\right) = \mathcal{N}\left(\sqrt{\overline{\alpha_k}}x_0, \left(1-\overline{\alpha_k}\right)I\right)$, with $\overline{\alpha}_k=\prod_{i=1}^k\left(1-\beta_i\right)$.
Similarly, the \emph{backwards} process is also modeled as a Markov chain with $\mathcal{K}$ steps and is defined as:  
\begin{equation} 
\label{eq:reverse_process}
p_\theta(x_{0:\mathcal{K}}) = p(x_\mathcal{K})\prod_{t=1}^{\mathcal{K}} p_\theta(x_{k-1} \mid x_k), \text{ with } x_\mathcal{K}\sim\mathcal{N}\left(0, I\right).
\end{equation} 
Starting from pure random noise, a data sample is generated by iteratively sampling from the reverse transition kernel $p_{\theta}\left(x_{k-1}\mid x_{k}\right)$ until $k=1$. The learnable transition kernel introduced in \cite{hoDDPM} takes the form:
\begin{equation}
    \label{backwardsTransition}
    p_{\theta}\left(x_{k-1} \mid x_k \right) = \mathcal{N}\left( \mu_{\theta}(x_k, k), \sigma_\theta\left(x_k, k\right)I\right),
\end{equation}
where the variance $\sigma_{\theta}\left(x_k, k\right)$ is uniquely determined by $\beta_k$. The mean parameter $\mu_{\theta}\left(x_k, k\right)$ additionaly also depends on a learnable denoising function $\epsilon_{\theta}\left(x_k, k\right)$, usually in the form of a neural network, that estimates the noise added to $x_k$.
With their work on Conditional Score-based Diffusion Models for Probabilistic Time Series Imputation (CSDI) \cite{tashiro2021csdi}, Tashiro et al. adapted DDPMs specifically for the task of time-series imputation. Let $x\in \mathbb{R}^{D\times L}$ be a multivariate time-series with missing values, where $D$ and $L$ represent the numbers of channels and of timesteps, respectively. It is possible to identify the missing values of $x$ using a binary mask $M\in\{0,1\}^{D\times L}$, where $M_{i,j}=0$ if the corresponding element of $x$ is missing. The task of time-series imputation focuses on estimating these missing values given the remaining observed values of the time-series. CSDI adapts equations (\ref{eq:reverse_process}) and (\ref{backwardsTransition}) to perform conditional imputation, taking form:
\begin{equation*}
\label{backwardsCSDI}
    \begin{aligned}   
    &p_\theta\left(x_{0:\mathcal{K}}^{mis}\mid x_{0}^{obs}\right) = p\left(x_\mathcal{K}^{mis}\right)\prod_{t=1}^{T}p_{\theta}\left(x_{k-1}^{mis}\mid x_k^{mis}, x_0^{obs}\right), \text{where } x_\mathcal{K}^{mis}\sim \mathcal{N}\left(0, I\right) \\
&p_\theta\left(x_{k-1}^{mis}\mid x_{k}^{mis}, x_0^{obs}\right) = \mathcal{N}\left(\mu_{\theta}\left(x_k^{mis}, k \mid x_0^{obs}\right), \sigma_\theta\left(x_k^{mis},k \mid x_0^{obs}\right)I\right) ,
    \end{aligned}
\end{equation*}
where $x^{mis}$ represents missing values of the time series for which an imputation is necessary, while $x^{obs}$ represents the observed parts. By iteratively sampling from the reverse conditional transition kernel $p_\theta\left(x_{k-1}^{mis}\mid x_{k}^{mis}, x_0^{obs}\right)$ for all steps $k=\mathcal{K},\dots,1$, CSDI leverages information from observed time points to generate plausible imputations for the missing values of the series.
\subsection{Bounded CopulaCPTS}
\label{app:copulaCPTS}

\begin{figure}[!t]     \centering
     \includegraphics[width=\textwidth]{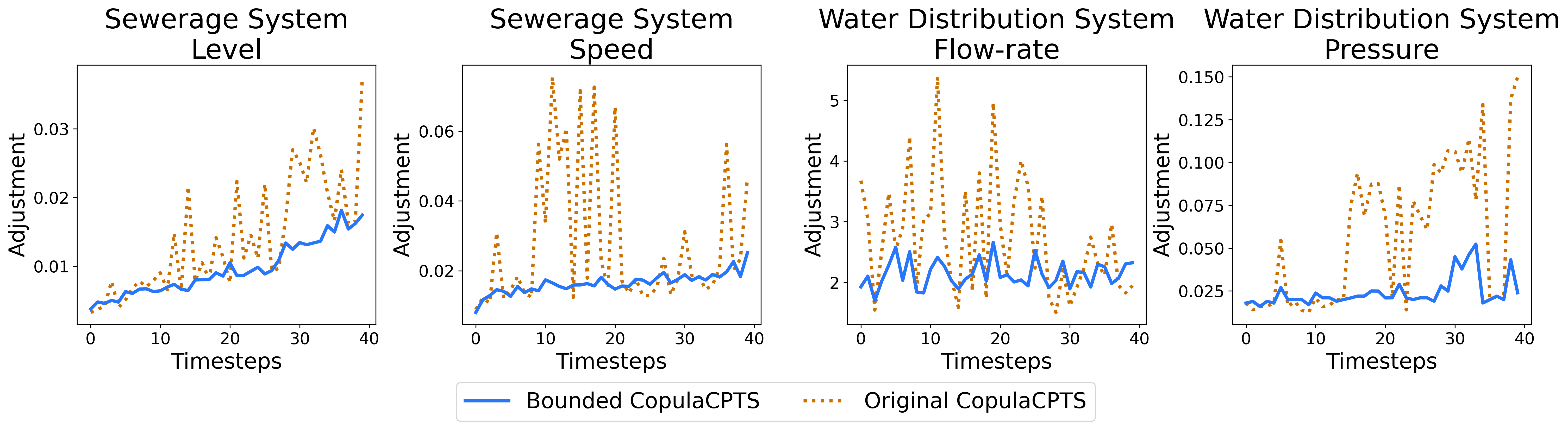}
     \caption{Comparison of the adjustment terms derived from the gradient-descent based version of CopulaCPTS (orange dotted) and our varaint which leverages bounded optimization (blue) for all four test cases.}
     \label{fig:boundedCopulaCPTS}
 \end{figure}

We propose a novel implementation of the CopulaCPTS method \cite{sun2023copulaCPTS} in which the gradient-descent optimization algorithm is replaced with a bounded optimization approach. Our proposal explicitly encodes efficiency as the objective to maximize, while still imposing validity constraints. Fig.~\ref{fig:boundedCopulaCPTS} shows a qualitative evaluation demonstrating this result, as the adjustment terms we derive (blue line) are often smaller and present less variability than CopulaCPTS's original output (orange dotted line).
\begin{table}[!t]
    \centering
    \caption{Comparison of coverage rates, adjustment average and standard deviation between the original CopulaCPTS and our bounded implementation. Our implementation achieves similar coverage rates to the original, while only requiring smaller and less variable adjustments.}
\label{tab:copulaCPTS}
\adjustbox{max width=\linewidth}{
\begin{tabular}{l|ccc|ccc|}
                      & \multicolumn{3}{c|}{{\color[HTML]{CC7000} \textbf{Original CopulaCPTS}}}                     & \multicolumn{3}{c|}{{\color[HTML]{2877FA} \textbf{Bounded CopulaCPTS}}}                     \\ \cline{2-7} 
\multicolumn{1}{c|}{} & \multicolumn{1}{c|}{Coverage (\%)} & \multicolumn{1}{c|}{Avg. Adjustment} & Std. Adjustments & \multicolumn{1}{c|}{Coverage (\%)} & \multicolumn{1}{c|}{Avg. Adjustment} & Std. Adjustment \\ \hline
Sew. - Level          & \multicolumn{1}{c|}{89.93}         & \multicolumn{1}{c|}{0.014}           & 0.008            & \multicolumn{1}{c|}{90.41}         & \multicolumn{1}{c|}{0.010}           & 0.004           \\
Sew. - Speed          & \multicolumn{1}{c|}{90.45}         & \multicolumn{1}{c|}{0.028}           & 0.020            & \multicolumn{1}{c|}{89.97}         & \multicolumn{1}{c|}{0.016}           & 0.003           \\
WDS - Flow-rate       & \multicolumn{1}{c|}{90.23}         & \multicolumn{1}{c|}{2.714}           & 0.942            & \multicolumn{1}{c|}{89.69}         & \multicolumn{1}{c|}{2.140}           & 0.219           \\
WDS - Pressure        & \multicolumn{1}{c|}{89.89}         & \multicolumn{1}{c|}{0.057}           & 0.042            & \multicolumn{1}{c|}{90.22}         & \multicolumn{1}{c|}{0.024}           & 0.008           \\ \hline
\end{tabular}
}
\end{table}
 
Additionally, Table~\ref{tab:copulaCPTS} shows a quantitative comparison of the adjustment terms provided by both methods over the four considered datasets. Both methods achieve similar levels of coverage, but our bounded implementation produces adjustment terms which are both smaller on average and present less variability. This improvement is consistent across all the evaluated datasets and results in conformalized predictive intervals which are smoother, offering a more robust starting point for the anomaly detection phase of CoCAI.

\section{Additional Plots}\label{app:plots}
\renewcommand{\thefigure}{B\arabic{figure}}
\setcounter{figure}{0}
This section contains a collection of experimental results for each of the four test cases examined in Section~\ref{sec:experiments}. For each scenario we include various combinations of anomaly scores and interval coverages.
\begin{figure}[!t]     \centering
     \includegraphics[width=\textwidth]{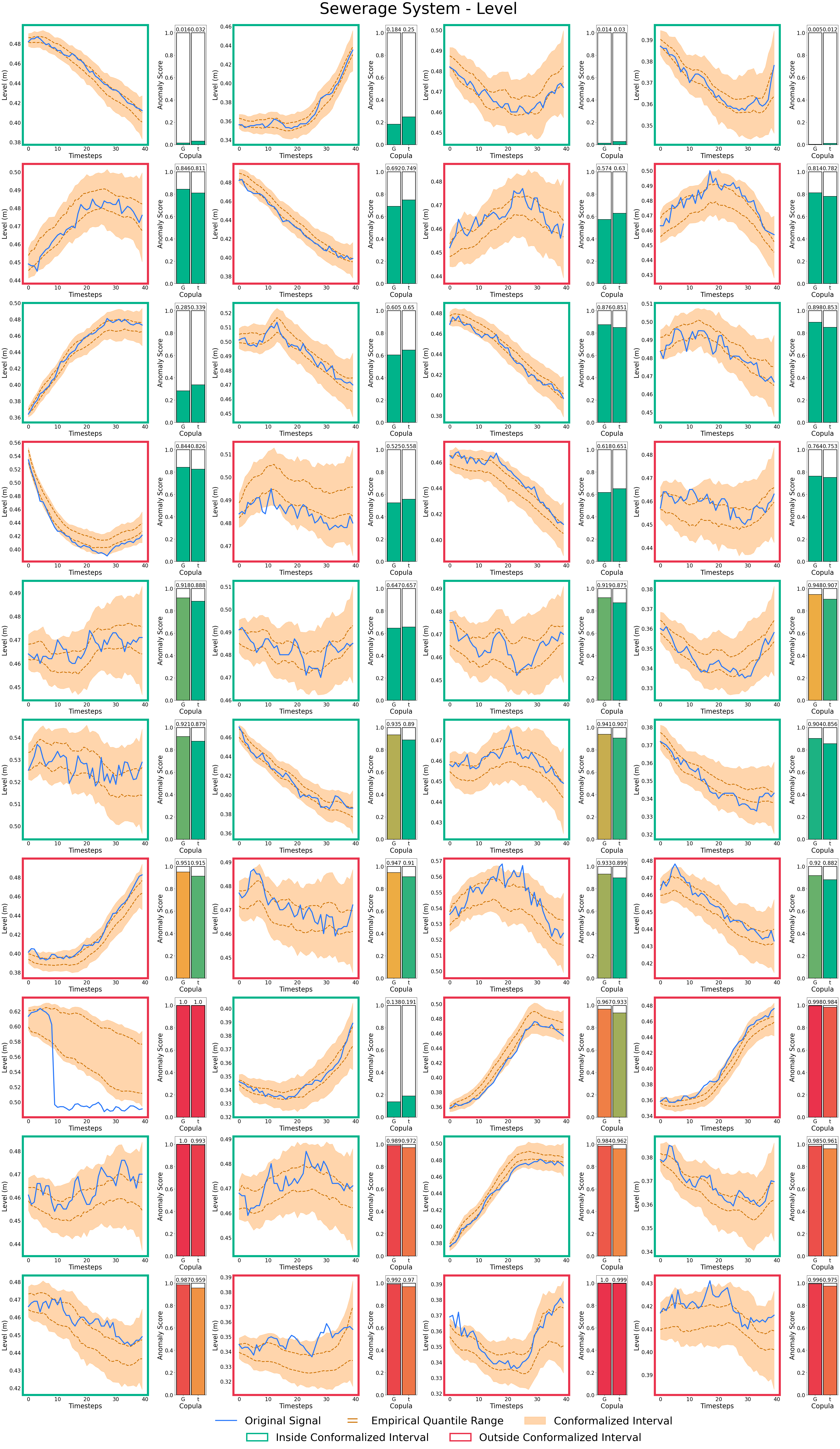}
     \vspace{-.5cm}
     \caption{A collection of prediction and anomaly detection results for various sewerage system \emph{level} time series, covering different combinations of anomaly scores and interval coverages.}
     \label{fig:result_plot_SEW_00}\vspace{-.5cm}
 \end{figure}

\begin{figure}[!t]     \centering
     \includegraphics[width=\textwidth]{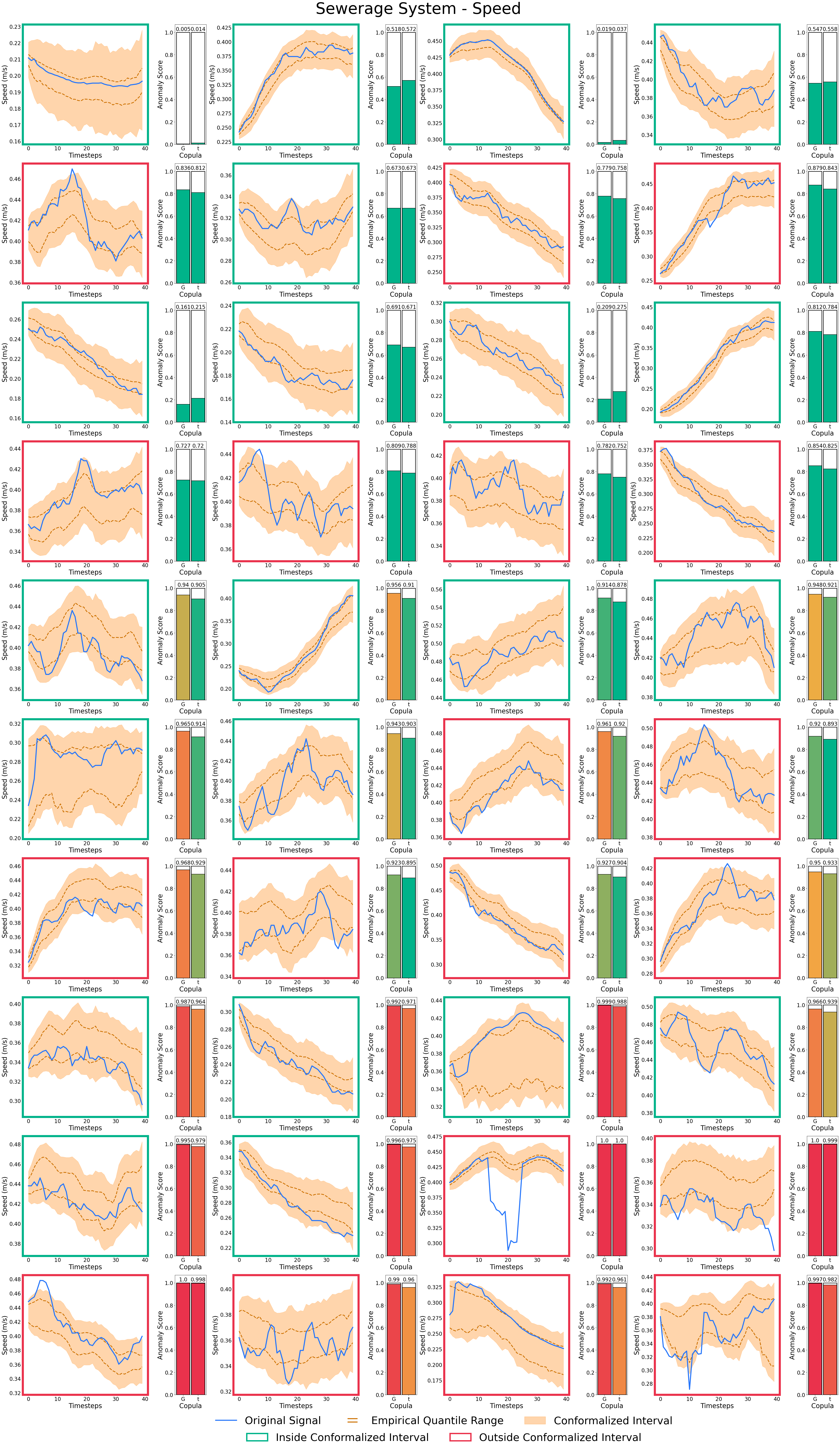}
     \vspace{-.5cm}
     \caption{A collection of prediction and anomaly detection results for various sewerage system \emph{speed} time series, covering different combinations of anomaly scores and interval coverages.}
     \label{fig:result_plot_SEW_01}\vspace{-.5cm}
 \end{figure}

\begin{figure}[!t]     \centering
     \includegraphics[width=\textwidth]{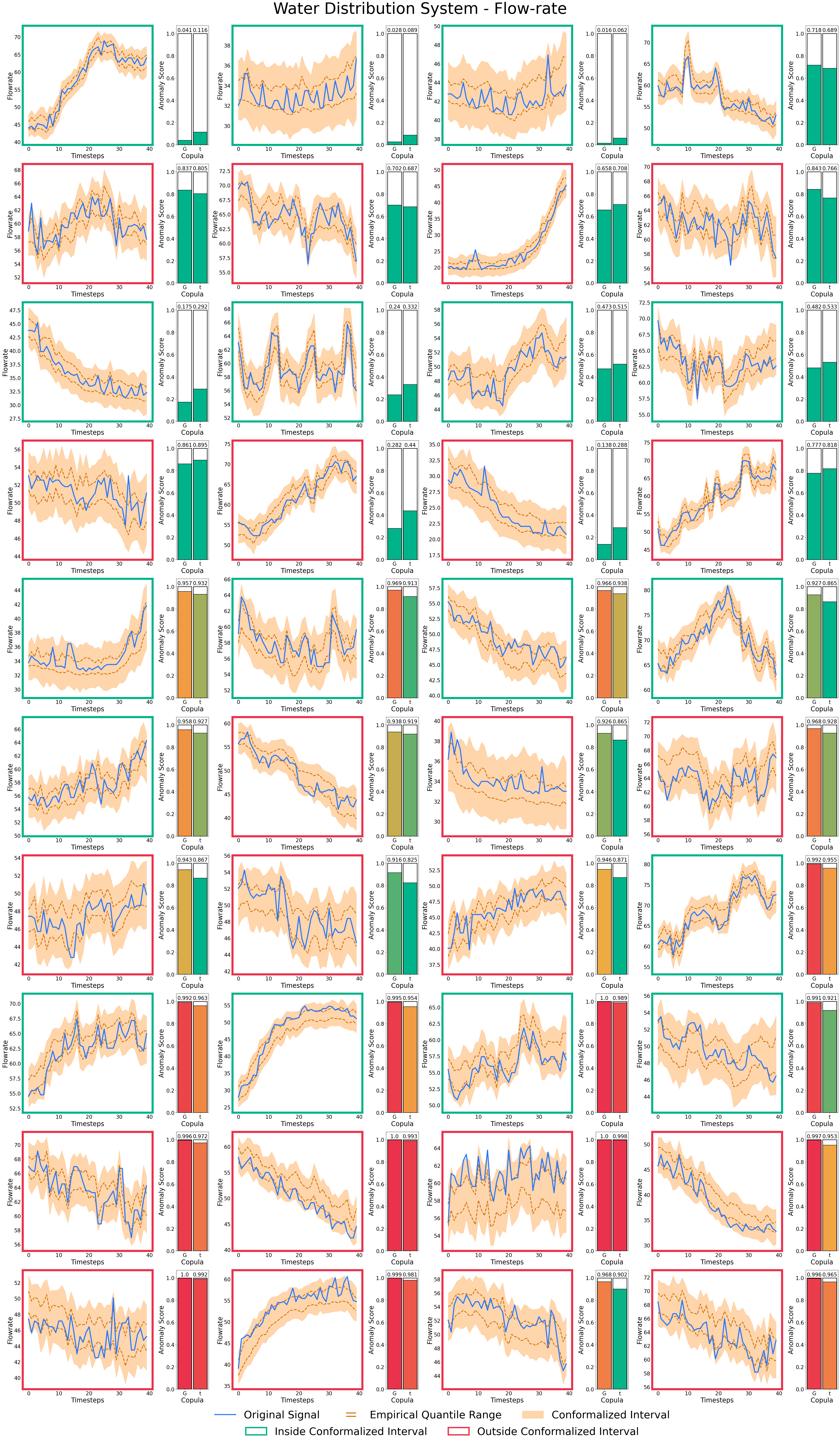}
     \vspace{-.5cm}
     \caption{A collection of prediction and anomaly detection results for various WDS water \emph{flow-rate} time series, covering different combinations of anomaly scores and interval coverages.}
     \label{fig:result_plot_WDS_02}\vspace{-.5cm}
 \end{figure}

  \begin{figure}[!t]     \centering
     \includegraphics[width=\textwidth]{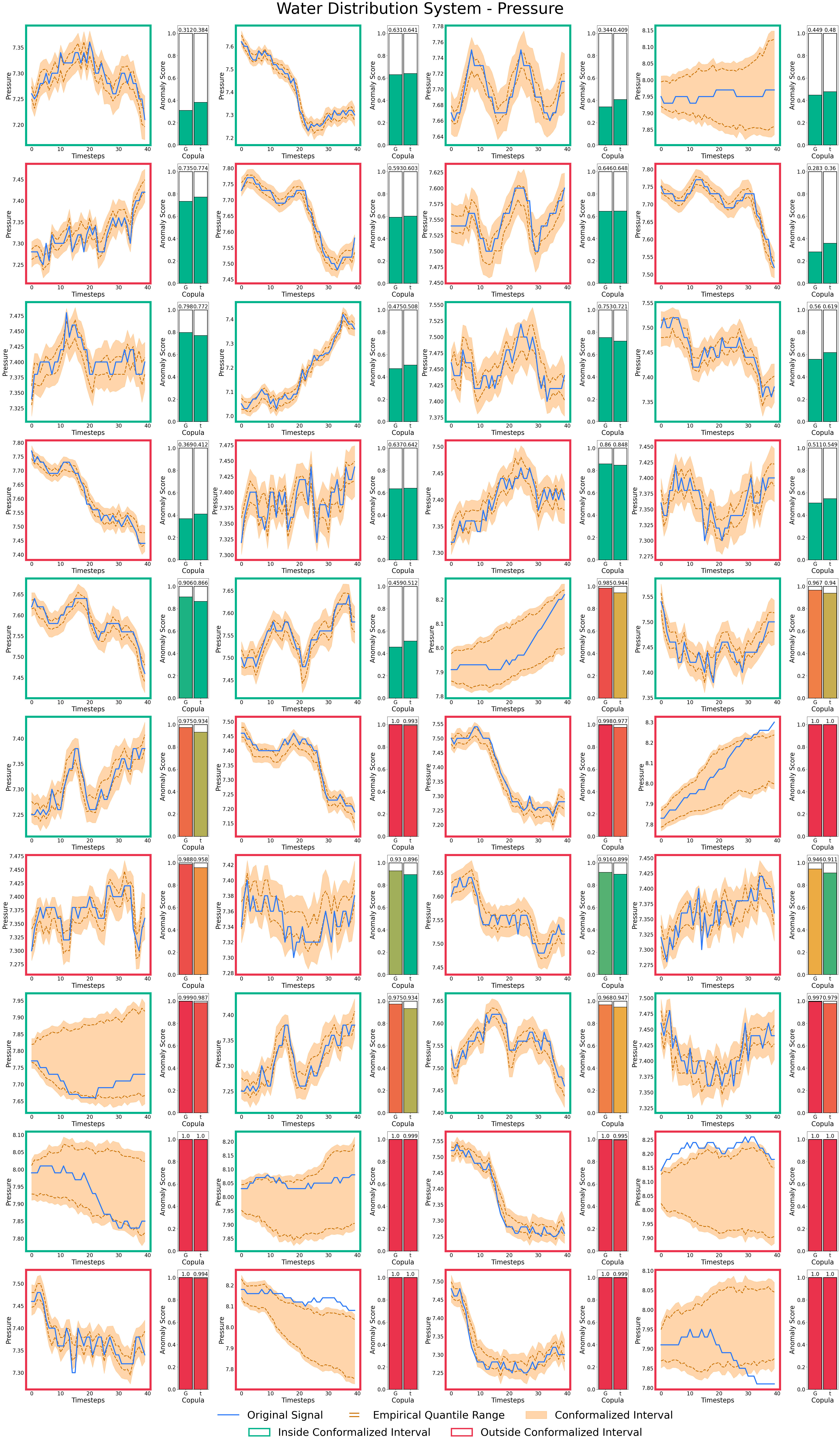}
     \vspace{-.5cm}
     \caption{A collection of prediction and anomaly detection results for various WDS \emph{pressure} time series, covering different combinations of anomaly scores and interval coverages.}
     \label{fig:result_plot_WDS_00}\vspace{-.5cm}
 \end{figure}

\end{document}